\documentclass{article}
\usepackage[T1]{fontenc}
\usepackage{iclr2027_conference,times}
\usepackage{amsmath,amssymb,amsfonts}
\usepackage{amsthm}
\newtheorem{proposition}{Proposition}
\newtheorem{remark}{Remark}
\usepackage{booktabs}
\usepackage{longtable}
\usepackage{tabularx}
\usepackage{array}
\usepackage{needspace}
\usepackage{graphicx}
\usepackage{tikz}
\usetikzlibrary{arrows.meta}
\usepackage{xcolor}
\usepackage{multirow}
\usepackage{placeins}

\usepackage[hidelinks]{hyperref}
\usepackage{url}

\graphicspath{{figures/}}
\definecolor{figblue}{HTML}{2a78d6}
\definecolor{figorange}{HTML}{eb6834}
\definecolor{figgray}{HTML}{52514e}

\newcommand{\CGCov}{1{,}920/1{,}920}
\newcommand{\ObsCov}{68/68}

\newcommand{\TorchNullAcc}{83}
\newcommand{\TorchNullNll}{0}
\newcommand{\TorchNullLast}{0}
\newcommand{\TorchGaussAcc}{80}
\newcommand{\TorchGaussNll}{0}

\newcommand{\TorchPosLowAcc}{47}
\newcommand{\TorchPosLowNll}{1}

\newcommand{\TorchPosHighAcc}{46}
\newcommand{\TorchPosHighNll}{5}

\newcommand{\QNegSeedTests}{72}
\newcommand{\QNegSeedQone}{10}
\newcommand{\QNegSeedQtwo}{8}
\newcommand{\QNegSeedQthree}{10}
\newcommand{\QPosSeedTests}{42}
\newcommand{\QPosSeedQone}{28}
\newcommand{\QPosSeedQtwo}{24}
\newcommand{\QPosSeedQthree}{24}

\newcommand{\QUnits}{103}
\newcommand{\QOneAccMed}{0.44}
\newcommand{\QTwoAccMed}{0.32}
\newcommand{\QThreeAccMed}{0.19}
\newcommand{\QThreeAccMin}{0.025}
\newcommand{\QTwoAccMin}{0.126}
\newcommand{\QOneAccMin}{0.277}

\newcommand{\ObsResultsSentence}{Recomputing all 272 observed-label MLP tests under log-loss selection
changes the seed-level conditional flags from 19 to 21, and only nine tests are flagged under both policies; the
two MLP R1 cells become a different two, AR-block/DeiT/C-100 at $O_0$ replacing AR-full/DeiT/C-100
at $O_2$ (Table~\ref{tab:nllobserved}). The ten linear R1 cells are unaffected, because the linear
probe involves no validation-based selection.}

\newcommand{\SHundredExceed}{9}
\newcommand{\SHundredMaxSteps}{12}

\newcommand{\SHundredBelowFiftyWord}{three}

\newcommand{\SHundredOldExceed}{6}

\newcommand{\CapFigSetup}{%
\textbf{Setup of the audit.} In one forward pass a trained vision transformer $f$ maps an input $x$
to an output view $O_j$ ($j=0,\dots,3$) and to $R\in\mathbb{R}^{L_f}$, its per-layer routing
entropies. A logistic generator $g(O_0)$ is fitted to real correctness on half $A$ and frozen; on
the disjoint half $B_{\rm test}$, $T^*_i\sim\mathrm{Bernoulli}\{g(O_{0,i})\}$. Every view determines
$O_0$ and $R$ is not used for the labels, so $T^*\perp R\mid O_j,A$. Probes with the same
architecture and input width are cross-fitted on $B_{\rm test}$ to predict $T^*$ from $(O_j,0)$ and from $(O_j,R)$
and scored by held-out log loss or Brier score $\widehat L_S$; the RAW gain $\widehat\Delta_{\rm raw}$
of Eq.~\eqref{eq:contrasts} is positive when routing lowers the loss. The families differ in the
reference for the probe on $(O_j,R)$: the probe on $(O_j,0)$ (RAW); the same probe refitted after a
uniform (GLOBAL) or output-matched (MATCHED) permutation $\pi$ of routing rows; or refits on 79
permutations drawn with weights $\prod_i\hat q_j(R_{\pi(i)}\mid O_{j,i})$, with $\hat q_j$ fitted on $A$
(CPT, Eqs.~\eqref{eq:cpt}--\eqref{eq:pvalue}). The input image is illustrative.}
\newcommand{\CapFigTaxonomy}{%
\textbf{Large probe gains need not survive routing-specific controls.}
All 22 families use the same MLP and $O_0$; points are mean held-out log-loss contrasts
across training seeds, with positive values favouring real routing.
Families are ordered by mechanism, backbone and dataset, independently of outcomes.
The full screen requires both scores and seed replication: 0/22 families pass across
all views and both probes, which does not establish absence of incremental information.}
\newcommand{\CapFigMech}{%
\textbf{Probe comparisons under the exact label null of Fig.~\ref{fig:setup}, and the role of
checkpoint selection.}
\emph{A:} percentages flagged by RAW, GLOBAL and MATCHED (both 95\% bootstrap intervals above zero)
and by CPT (both $p$-values at most $0.025$). MLP results use accuracy-based checkpoint selection.
600 evaluations per probe at $O_0$ and $O_1$ and 360 at $O_2$ and $O_3$, sharing 600 label vectors.
\emph{B:} the same six models at $O_0$, 20 label draws each: every MLP fit runs for 100 epochs and
its trajectory is read at the checkpoint with the best validation accuracy and at the one with the
lowest validation log loss, in the stock scikit-learn estimator and in an independent PyTorch
implementation. Right: mean held-out loss of the output-only probe minus the frozen generator's
loss on the same label draws, a diagnostic rather than a population excess risk
(Tables~\ref{tab:metric} and~\ref{tab:torchsel}).}

\newcommand{\CapFigHazards}{%
\textbf{Two control hazards motivate separate representation and assignment checks.}
\emph{A:} a single Swin/AR-block/C10 seed-0 MLP compares each appended block with its own
width-matched padded baseline (routing width 8, output-extra width 4); intervals are
paired 95\% bootstrap intervals.
\emph{B:} token-averaged routing weights are compared with noise and GLOBAL controls;
labels count positive log-loss intervals among three seeds, not complete two-score passes.
Both panels are descriptive; neither identifies optimisation error as the sole cause.}
\newcommand{\CapFigValidity}{%
\textbf{Historical component-level power does not establish CPT family-rule power.}
RAW and GLOBAL curves show two routing widths and three sample sizes from the original
synthetic benchmark, with 200 repeats per point; the full sample-size grid remains in
App.~\ref{app:power}.
The independent-split linear CPT endpoint is separate (App.~\ref{app:cpt}); neither
component curves nor that endpoint establish conjunction power.}

\title{Routing Probes Can Improve Without\\New Information: An Exact-Null Audit of\\Uncertainty Beyond Model Outputs}

\author{\begin{tabular}{@{}p{0.30\textwidth}@{}p{0.30\textwidth}@{}p{0.30\textwidth}@{}}
Wenhao Liang & Lin Yue & Wei Emma Zhang \\[8pt]
Mingyu Guo & Olaf Maennel & Weitong Chen
\end{tabular}}

\iclrfinalcopy
\begin{document}
\maketitle
\lhead{Preprint.}
\begin{abstract}
Routing signals of modern vision transformers---expert gates, attention-residual weights and
halting scores---often improve probes that predict whether the model is correct, and the
improvement is commonly read as evidence that routing carries information about errors beyond the
model's outputs. We test this inference directly: keeping real output--routing pairs, we redraw
correctness labels from a frozen output-only generator fitted on disjoint data, so that routing is
uninformative by construction. Under this exact label null, a width-matched MLP comparison still
reports a routing gain in 51.3\% of confidence-only evaluations (308/600), while a linear
comparison reports none. Holding each training trajectory fixed on a six-model panel and selecting
the checkpoint by validation log loss instead of validation accuracy removes the detections
(50/120 to 0/120, and \TorchNullAcc/120 to \TorchNullNll/120 in an independently implemented
probe), identifying accuracy-based checkpoint selection as the cause; across all output views the
raw detection rate falls from 27.5\% (528/1,920) to zero observed detections. The repaired
comparison is not sensitive, detecting an implanted signal of about 0.005 nats in 0/20 replicates
in each of two matched settings, whereas a conditional permutation test built on an estimated
routing law detects it in 11/20 and 10/20 and rejects rarely under the null. On real correctness
labels, the conditional analysis yields model-relative evidence in five DeiT attention-residual
families; in four it persists under two specified variants of the conditional law, and no family
passes an additional noise criterion. Fitting a better probe and testing for incremental
information are different problems, and each needs its own validation.
\end{abstract}

\section{Introduction}
\label{sec:intro}

Expert gates, halting scores and attention-residual weights are computed during inference and
correlate with errors. Probes that read them alongside a model's confidence can predict
correctness better than probes that read the confidence alone, and gate entropy has been used
in this way for segmentation uncertainty and out-of-distribution detection
\citep{pavlitska2025extracting,lee2026unseen}. This paper examines a further inference: from
``the probe improved'' to ``routing carries information about errors that the outputs do not''.

That inference does not follow from the improvement itself. If correctness depends on $O^2$,
adding $R=O^2$ helps a linear probe although $O$ already determines $R$; Proposition~\ref{prop:gap}
shows that the gap persists with unlimited data and equal parameter counts after zero-padding.
This is the distinction between usable information and information content
\citep{xu2020usable,hewitt2021conditional}. What is less clear is how often, and for what
reasons, the inference fails for the probes, sample sizes and controls used in practice, where
routing and outputs are strongly dependent.

We measure this with a null whose answer is known. For each of six trained vision transformers
we keep the real output--routing pairs but redraw correctness labels from a logistic function of
the model's confidence, fitted on a disjoint half of the data and then frozen
(Fig.~\ref{fig:setup}). Routing then carries no information beyond the outputs. A width-matched MLP probe nevertheless reports a
routing gain in 308 of 600 evaluations at the confidence-only view; the linear probe reports
none. Across all four output views, the MLP comparisons against globally and output-matched
shuffled routing, which are intended as controls, report gains in 308 and 293 of 1,920
evaluations (\S\ref{sec:exactnull}).

\begin{figure}[t]
\centering
\input{sections/fig_setup}
\caption{\CapFigSetup}\label{fig:setup}
\end{figure}

The MLP is scikit-learn's \texttt{MLPClassifier} with early stopping enabled, which selects its
checkpoint by validation accuracy while every comparison is scored by log loss and Brier score.
To isolate this choice we train each probe once for a fixed number of epochs and read the same
trajectory at the checkpoint preferred by validation accuracy and at the one preferred by
validation log loss. Raw detections fall from 50/120 to 0/120. An independently implemented
PyTorch probe shows the same contrast (\TorchNullAcc/120 against \TorchNullNll/120), as does
replacing routing by independent Gaussian features (\TorchGaussAcc/120 against
\TorchGaussNll/120), so the failure is not specific to this estimator and does not require
routing structure. Applying log-loss selection to all 1,920 MLP evaluations of the
null benchmark removes all 528 raw detections and most shuffle-control detections
(\S\ref{sec:mechanism}).

Removing detections under the null does not make the comparison sensitive. In two matched
synthetic settings with an implanted signal of about 0.005 nats, the repaired raw comparison
detects it in 0/20 replicates in each and its mean fitted gain stays negative, whereas a
conditional permutation test (CPT) that refits the probe on assignments drawn from an estimated
conditional law of routing given outputs detects it in 11/20 and 10/20
\citep{berrett2020conditional}. With the true
conditional law such a test has guaranteed level; with an estimated law its behaviour must be
measured, and on the null benchmark it rejects in 59 of 3,840 evaluations.

Applied to real correctness labels across 22 routing families and 68 training runs, the
conditional test finds replicated conditional-assignment evidence (R1) in 12 of 176
family--probe--view cells, all in five DeiT attention-residual families, and none of these passes
an additional same-width noise criterion (R2). Eight of the 12 cells keep R1 under two specified
variants of the conditional model, both of which fit routing less well; in the other four the
number of positive seeds falls from two to one.

Our contributions are three empirical results. We measure how raw and shuffle-based probe
comparisons behave under an exact label null that preserves real output--routing dependence. We
isolate the effect of checkpoint selection on fixed training trajectories, in two probe
implementations, and evaluate a log-loss policy across the complete MLP null benchmark. We then
measure the sensitivity of the corrected comparison against a conditional reference and report
the resulting model-relative evidence on real labels.

\section{Incremental information and probe utility}
\label{sec:formal}

\paragraph{Population target.}
Fix a trained classifier and an evaluation distribution. Let $T\in\{0,1\}$ indicate top-1
correctness, and let $O$ and $R$ denote specified output and routing representations. For a
strictly proper loss $S$ with finite Bayes risks, define
\begin{equation}
\label{eq:gain}
\mathrm{Gain}_S(R\mid O)=\mathcal R_S^\star(O)-\mathcal R_S^\star(O,R),
\end{equation}
where $\mathcal R_S^\star(X)=\inf_f\mathbb E[S(f(X),T)]$ is the Bayes risk over measurable
predictors $f$ of $T$ from $X$. Writing $\eta_O=P(T=1\mid O)$ and $\eta_{OR}=P(T=1\mid O,R)$,
strict propriety gives $\mathrm{Gain}_S\ge0$, with equality exactly when $\eta_{OR}=\eta_O$
almost surely, that is, when $T\perp R\mid O$. Log loss yields $I(T;R\mid O)$ in nats and squared
loss yields $\mathbb E[(\eta_{OR}-\eta_O)^2]$ \citep{gneiting2007proper}. For any deterministic
summary $\phi$, $I(T;\phi(R)\mid O)\le I(T;R\mid O)$. A null result for the routing entropies used
here therefore does not rule out information in the full routing trace, whereas positive
information in the entropies would imply it for the trace.

\paragraph{Output representations.}
We evaluate four views: $O_0$, top-class confidence after a clipped logit transform;
$O_1$, confidence, predictive entropy, top-two margin, logit norm and runner-up probability;
$O_2$, sorted class probabilities; and $O_3$, sorted logits. The last two keep the shape of the
scores but remove class identity, so they cannot test sufficiency of the complete,
class-labelled output. The views are not nested: $O_1$ includes the logit norm, which $O_2$ cannot
recover (App.~\ref{app:protocol}). Each view defines its own conditional question.

\paragraph{Why a probe gain can be positive under the null.}
A fitted probe searches a restricted function class, whereas Eq.~\eqref{eq:gain} compares
optimal predictors over all measurable functions. The distinction persists at the population
level.

\begin{proposition}[Positive probe gain under conditional sufficiency]
\label{prop:gap}
There exist a joint law for $(T,O)$ and a deterministic $R=\phi(O)$ such that
$T\perp R\mid O$, but the logistic classes
$\mathcal F_O=\{\sigma(aO+b)\}$ and
$\mathcal F_{O,R}=\{\sigma(aO+cR+b)\}$ satisfy
\[
\inf_{f\in\mathcal F_O}\mathcal R_{\log}(f)
-\inf_{f\in\mathcal F_{O,R}}\mathcal R_{\log}(f)=I(T;O)>0.
\]
\end{proposition}

Take $O$ uniform on $\{-1,0,1\}$, $R=O^2$ and
$T\mid O\sim\mathrm{Bernoulli}(\sigma(O^2))$. The augmented class contains the Bayes predictor.
Symmetry and convexity make the best output-only logistic predictor constant, giving a gain of
$H(\mathbb E\eta_O)-\mathbb E H(\eta_O)\approx0.0257$ nats, although $R$ is a function of $O$ and
the incremental information is zero. App.~\ref{app:proof} gives the proof and a noisy-routing
extension. Zero-padding leaves the output-only class unchanged, so equal nominal parameter
counts do not equalise what the two probes can express as functions of $O$.

A probe gain is therefore not sufficient to infer information. The benchmark below exercises a
different route to the same error. At $O_0$ the label generator is a logistic function of $O_0$,
so the output-only logistic class contains it and Proposition~\ref{prop:gap} cannot explain a gain
there. The MLP probes being compared share an architecture and a nominal input width, but the
checkpoints selected for them can leave different fitting errors and hence different held-out
losses. A richer baseline class
addresses approximation error; it does not address checkpoint selection, and a better-fitted
comparison is still a predictive-utility contrast rather than a test of incremental information.

\section{The Routing Information Audit}
\label{sec:ria}

\subsection{Matched probing and the screening rule}
\label{sec:probing}

The routing representation is the per-layer entropy profile; a twelve-run panel also tests raw
routing weights (App.~\ref{app:alpha}). At each output view we fit probes to four feature sets:
zero-padded $O_j$, $(O_j,R)$, $(O_j,\pi(R))$ and $(O_j,\varepsilon)$, where $\pi$ shuffles
routing rows and $\varepsilon$ is an independent Gaussian block of the same width. The shuffle
keeps the empirical distribution of routing vectors but breaks both the example assignment and
the dependence on $O_j$, so it is an assignment control rather than a draw from the conditional
null.

The probes are logistic regression and a one-hidden-layer MLP with 64 units, each with fixed
hyperparameters for every feature set. The MLP is scikit-learn 1.6.1's \texttt{MLPClassifier}
with \texttt{early\_stopping=True}, which holds out 10\% of each training fold, scores it by
validation accuracy after every epoch, stops after 15 epochs without improvement and restores
the weights with the best validation accuracy. Its training loss remains the regularised log
loss; only the validation criterion concerns accuracy. Logistic regression has no validation-based
selection. Features are standardised on training folds, predictions are cross-fitted over five
folds with two repeats and averaged over repeats, and log-loss and Brier differences are
bootstrapped with 2,000 paired example resamples. The resamples hold the fitted out-of-fold
predictions fixed, so the intervals do not include variability from refitting the probes.

With $\widehat L_S(X)$ the held-out loss of the probe using $X$, the three contrasts are
\begin{align}
\widehat\Delta_{\rm raw}&=\widehat L_S(O_j,0)-\widehat L_S(O_j,R),\nonumber\\
\widehat\Delta_{\rm shuffle}&=\widehat L_S(O_j,\pi(R))-\widehat L_S(O_j,R),\label{eq:contrasts}\\
\widehat\Delta_{\rm noise}&=\widehat L_S(O_j,\varepsilon)-\widehat L_S(O_j,R),\nonumber
\end{align}
with positive values favouring real routing. In the label-null benchmarks a comparison is
\emph{flagged} when the 95\% bootstrap intervals for both scores lie above zero in the same
replicate. This screening rule concerns the realised fitted contrast; it has no proven level for
the incremental-information null. The pre-specified screen applied to real labels requires a
component to have a positive interval in at least two training seeds for each score (the seeds
may differ between scores) and requires the raw, shuffle and noise components to pass for the
same probe and view; App.~\ref{app:protocol} defines it and its Type~0--III labels, Type~0 meaning
that no view passes. A post-hoc output-matched shuffle (MATCHED, App.~\ref{app:matched}) restricts
reassignment to local neighbourhoods in $O_j$; it is an approximate assignment control outside
the pre-specified screen.

\subsection{A conditional reference for the probe statistic}
\label{sec:cptmethod}

Because the global shuffle breaks the dependence of routing on the outputs, we also compare the
observed raw contrast with contrasts computed on routing reassigned under an estimated
conditional law $\hat q_j(R\mid O_j)$ \citep{berrett2020conditional}. A fixed split gives disjoint
halves: 5,000 examples in A for selecting and fitting $\hat q_j$, and 5,000 in $B_{\rm test}$ for
every probe and statistic. No correctness labels enter $\hat q_j$. The working model maps routing
coordinates to normal scores by rank, predicts them from standardised $O_j$ by ridge regression
and uses a shared residual correlation; it is the tier selected by held-out conditional
log-likelihood on A in all 272 run--view fits (App.~\ref{app:cpt}). Routing lies in
$\mathbb R^{L_f}$ with family-specific $L_f\in\{8,12,16,22,23\}$. The assignment weights are
\begin{equation}
\label{eq:cpt}
P_{\hat q_j}(\pi\mid O_j,\{R_i\})\ \propto\ \prod_i\hat q_j(R_{\pi(i)}\mid O_{j,i}).
\end{equation}
A reversible hub-and-spoke sampler draws $B=79$ reference assignments with $S=50$ transition
rounds per leg. Each assignment passes through the same five-fold, two-repeat fitting and
selection procedure as the observed routing, with the same folds and seeds, and for each score
\begin{equation}
\label{eq:pvalue}
p_S=\frac{1+\sum_{b=1}^{79}\mathbf1\{\widehat\Delta_S^{(b)}\ge
\widehat\Delta_S^{\rm obs}\}}{80}.
\end{equation}
The grid has spacing $1/80$, so only $p_S\in\{0.0125,0.025\}$ meet the threshold $0.025$; in the
benchmarks a single conditional test is flagged when both of its $p$-values meet it. If
$\hat q_j$ were the true conditional law, the observed and reference statistics would be
exchangeable and this rank test would have level at most $0.025$ for any fixed fitting
procedure. Because $\hat q_j$ is estimated, its rejection rates are measured rather than
guaranteed.

\textbf{R1} requires both $p$-values to be at most $0.025$ in the same training seed, and at least
two such seeds in a family--probe--view cell; for the one family with five seeds this is two of
five. R1 is evidence that the observed assignment is unusual under $\hat q_j$, not an estimate of
$\mathrm{Gain}_S$. \textbf{R2} is an additional same-width robustness criterion: in the same seeds,
real routing must also beat the Gaussian block on the same $B_{\rm test}$ examples with positive
intervals for both scores, in at least two joint-positive seeds. R2 has no raw-gain or shuffle
conjunct, and beating a noise block is not necessary for conditional information.

\subsection{An exact label-null benchmark}
\label{sec:benchmark}

We fit one logistic generator $g(O_0)$ per model on A, freeze it and draw every $B_{\rm test}$ label
from it (Fig.~\ref{fig:setup}). Every evaluated $O_j$ determines $O_0$, so $T^*\perp R\mid O_j$ holds by construction
up to the numerical checks of App.~\ref{app:t4}. The benchmark uses six seed-0 models covering the
three routing mechanisms and three backbones. Each model contributes 100 label draws at
$O_0$ and $O_1$ and the first 60 of those draws at $O_2$ and $O_3$, giving each probe 600
evaluations at the first two views, 360 at the last two and 1,920 in total. Views and probes
reuse the same 600 label vectors, so pooled counts describe the benchmark; they are not
independent Bernoulli trials. The label null is exact given the generator; it does not make the
estimated-$q$ test exact.

\section{Results}
\label{sec:results}

The audit covers 22 families (routing variant, backbone and dataset): 18 attention-residual,
two mixture-of-experts (MoE) and two soft adaptive-depth (halting) families, built on DeiT-Small,
Swin-Tiny and ViT-B/16 and trained on CIFAR-10, CIFAR-100 and Tiny-ImageNet, with three training
seeds per family except DeiT/C-100/AR-full with five: 68 runs with 10,000 evaluation examples
each. The attention-residual checkpoints and evaluation records are those described by
\citet{liang2026routing} and are reused without retraining; the MoE and halting models are
specified in App.~\ref{app:config}. Configuration snapshots survive for 66 of the 68 runs and all
66 include a soft-binned calibration penalty; training quality varies across runs
(Table~\ref{tab:cachefacts}).

\subsection{What the comparisons report under the label null}
\label{sec:exactnull}

In the benchmark of \S\ref{sec:benchmark} routing carries no information beyond any evaluated
view, so every flag is a false positive for a claim of incremental information, even when the
routing-augmented probe really does predict the relabelled correctness better than the fitted
output-only probe. The MLP raw comparison flags 308 of its 600 evaluations at $O_0$ and 528 of
1,920 across views; the linear raw comparison flags none (Fig.~\ref{fig:mech}A, App.~\ref{app:t4}). The MLP
comparisons against globally and output-matched shuffled routing flag 308 and 293 of 1,920
evaluations, and the linear ones 17 and 12. The conditional test flags 30 MLP and 29 linear
evaluations, 59 of 3,840 in total. The failures are heterogeneous across models: at $O_0$ the MLP
raw comparison flags between 8 and 100 of the 100 draws per model (Table~\ref{tab:t4cells} lists
the conditional test by model).

\subsection{Checkpoint selection produces the raw detections}
\label{sec:mechanism}

At $O_0$ the generator is a logistic function of $O_0$, so the output-only probe class contains it
and Proposition~\ref{prop:gap} does not explain these flags. The MLP's validation criterion does.
Accuracy does not distinguish probability estimates that give the same hard predictions, so
checkpoints with similar validation accuracy can differ substantially in log loss, and the two
probes being compared need not reach their best accuracy at comparable points of training.

We instrument the stock scikit-learn estimator with a subclass that overrides only the method
scoring the internal validation split (App.~\ref{app:metric}), on six cells with 20 label draws
each. Arm~A returns the stock accuracy score and reproduces the stored estimator bit for bit
(57/120 flags). Arm~B returns the negative validation log loss, which changes both when training
stops and which checkpoint is kept, and gives 0/120. Because A and B differ in training length as
well as selection, two further arms fix the trajectory: every fit runs for 100 epochs and the same
recorded trajectory is read at the checkpoint with the best validation accuracy and at the one
with the lowest validation log loss. Flags fall from \textbf{50/120} to \textbf{0/120}. Averaged
over fits, accuracy selects epoch 18 for the output-only probe against 56 under log loss, and 37
against 42 for the routing-augmented probe, although every fit runs all 100 epochs. On the same
label draws the output-only probe's held-out loss exceeds that of the generator by 0.0264 nats
under accuracy selection and by 0.0007 under log-loss selection (Table~\ref{tab:metric}); the
apparent routing gain is largely this shortfall of the early-selected baseline.

The same design in an independently implemented PyTorch MLP (same width, penalty definition,
optimiser settings, folds, labels and scoring; different initialisation, arithmetic precision and
validation draw) gives \TorchNullAcc/120 flags at the accuracy-selected checkpoint and
\TorchNullNll/120 at the log-loss-selected one, and \TorchNullLast/120 at the final epoch
(Table~\ref{tab:torchsel}). Replacing routing by independent Gaussian features of the same width
leaves the pattern unchanged (\TorchGaussAcc/120 against \TorchGaussNll/120), so no routing
structure is needed. Which cells fail differs between the implementations, and the overall rate is
higher in PyTorch; the direction of the effect does not depend on the estimator. With an implanted
signal the accuracy-selected PyTorch comparison flags about as often as under the null
(\TorchPosLowAcc/60 and \TorchPosHighAcc/60 at about 0.005 and 0.01 nats), and the log-loss-selected
one flags \TorchPosLowNll/60 and \TorchPosHighNll/60, which anticipates the sensitivity result of
\S\ref{sec:power}.

Applying the log-loss stopping-and-selection policy of arm~B to all 1,920 MLP evaluations of the
benchmark removes all 528 raw detections and flags no new ones (Table~\ref{tab:nllcommon}). GLOBAL
and MATCHED detections fall from 308 to 27 and from 293 to 26. The fixed-trajectory result
explains the raw comparison, whose baseline is the zero-padded probe; GLOBAL and MATCHED compare
two augmented fits, so their decline shows that the same metric mismatch affects them, but their
checkpoint dynamics were not isolated separately. Conditional-test detections change from 30 to
22, yet only four evaluations are flagged under both policies: a small change in the aggregate
rate does not mean that individual decisions persist. None of the 1,920 dependent evaluations
flags a raw gain under log-loss selection, which does not make the probability of such a flag zero.

\begin{table}[b]\centering\small\setlength{\tabcolsep}{5pt}
\caption{All 1{,}920 common-generator MLP evaluations under both selection policies, paired on
identical labels, splits, permutation assignments and bootstrap draws. $n_{ab}$ counts
evaluations flagged by the accuracy policy ($a$) and by the log-loss policy ($b$); the
accuracy column remains the behaviour of the originally defined procedure. Linear results are
unchanged, because the linear probe has no validation-based selection, and are not shown. Zero detections in 1{,}920 dependent evaluations should not be read as
a zero population detection probability. Per-view rows: Table~\ref{tab:nllcommonfull}.}
\label{tab:nllcommon}
\begin{tabular}{lrrrrrrr}\toprule
Comparator & $n$ & $n_{00}$ & $n_{01}$ & $n_{10}$ & $n_{11}$ & acc (\%) & nll (\%) \\\midrule
RAW & 1920 & 1392 & 0 & 528 & 0 & 27.50 & 0.00 \\
GLOBAL & 1920 & 1598 & 14 & 295 & 13 & 16.04 & 1.41 \\
MATCHED & 1920 & 1617 & 10 & 277 & 16 & 15.26 & 1.35 \\
CPT & 1920 & 1872 & 18 & 26 & 4 & 1.56 & 1.15\\
\bottomrule\end{tabular}\end{table}

\subsection{Removing null detections does not create sensitivity}
\label{sec:power}

Sensitivity has to be measured separately. At view $O_1$ we implant a routing signal of known
strength into two fixed settings, DeiT/C-100/AR-block and Swin/C-10/AR-block, with disjoint halves
of 5,000 examples for the conditional model and for every statistic, and analyse the first 20
replicates of each under both MLP policies with shared labels, routing, folds, controls and
reference assignments (Table~\ref{tab:nllpanel}, App.~\ref{app:cpt}). Under log-loss selection the
raw comparison flags no null replicate, but it detects a signal of about 0.005 nats in 0/20
replicates in both settings and one of about 0.01 nats in at most 1/20. Its mean fitted gain stays
negative up to 0.005 nats in both settings, so the implanted signal rarely makes the fitted
contrast positive, let alone both of its intervals; these means describe the fitted probes and do
not separate estimation error from regularisation or optimisation effects.

The accuracy-selected comparison is no substitute. It flags 16/20 Swin replicates with signal but
also 11/20 without it (28/50 over all 50 null replicates of that setting), when the routing block
is independent Gaussian noise. The conditional test on the same problems detects the 0.005-nat
signal in 11/20 Swin/C-10 and 10/20 DeiT/C-100 replicates under log-loss selection (3/20 and 7/20
under accuracy selection) and flags 0/20 and 1/20 null replicates.

The wider benchmark gives the same picture for single tests (Table~\ref{tab:power}). The linear
conditional test rejects 1/200 and 4/200 null replicates and all replicates at a reference gain of
0.010 nats; adding the noise conjunct lowers detection at 0.005 nats from 191/200 and 200/200 to
133/200 and 162/200. The accuracy-selected MLP rows are weaker throughout. These are
single-replicate operating characteristics on fixed model outputs. They are not measurements of
the cross-seed R1 or R2 rules, whose power remains unestablished, and the reference gain of a
synthetic signal is not the same quantity as a conditional contrast observed on real labels.

Two separate properties explain why the conditional test behaves differently from the raw
comparison. The padded baseline enters every statistic of a test as the same constant $C$, so with
$\Delta=C-A$ for augmented loss $A$, $\Delta^{(b)}\ge\Delta_{\rm obs}$ exactly when
$A^{(b)}\le A_{\rm obs}$: a change to the baseline alone cannot change the $p$-value. Validity is a
different matter. With the true conditional law, applying one fixed fitting and selection
procedure to the observed and reference assignments makes the statistics exchangeable, whichever
selection rule is used; selection can then change power and individual decisions but not the
level bound. With an estimated law, rejection rates are empirical. The doubled-$S$ sampler check
of the conditional protocol, run on one cell per view (App.~\ref{app:cptdiagnostics}), moves
\SHundredExceed{} of 16 $p$-values by more than the specified one grid step (up to
\SHundredMaxSteps{} steps) and changes one of eight test decisions at $0.025$, so the check fails.

\subsection{Conditional-assignment evidence on real correctness labels}
\label{sec:cptresults}\label{sec:r1}\label{sec:r2}

On the real held-out labels R1 is positive in \textbf{12 of 176} family--probe--view cells, all in
five DeiT attention-residual families (AR-block and AR-full on C-10 and C-100, AR-full on
Tiny-ImageNet); ten cells are linear and two MLP, the MLP using the original accuracy-based
selection (Table~\ref{tab:t2r1}). Mean conditional contrasts
in these cells range from $+0.0012$ to $+0.0054$ nats, and some individual seeds are negative. R1
states that the observed routing assignment is unusual under $\hat q$; it is model-relative
evidence, not an estimate of $\mathrm{Gain}_S$. No cell meets R2 (\textbf{0/176}): of the 28
CPT-positive seed evaluations in R1 cells, three also beat noise on both scores, in different
cells. Failing R2 does not remove R1, because beating a Gaussian block is not necessary for
conditional information.

Working-model diagnostics are somewhat worse in the R1-positive groups (median Mahalanobis KS
0.171 against 0.126 elsewhere at $O_0$ and 0.180 against 0.144 at $O_2$, with overlapping ranges),
and in DeiT/C-100/AR-full the two seeds that supply the $O_0$ and $O_2$ rejections also have that
family's largest discrepancies (App.~\ref{app:cptdiagnostics}). Misspecification of $\hat q$ is
therefore an alternative explanation. In a post-hoc check with a protocol fixed before the runs
(Table~\ref{tab:qsens}, App.~\ref{app:qsens}), we recomputed every seed of the five families under
two fixed variants of the working model, each with hyperparameters tuned within A by three-fold
held-out conditional log-likelihood: a gradient-boosted mean (Q2), and the selected Q1 mean with an
output-dependent diagonal scale and a fitted shared residual correlation (Q3). Both fit routing
less well than Q1 in all 103 fits, and all three share the rank-normal Gaussian framework. The
Swin and ViT cells with the same dataset, routing variant, probe and view were recomputed as a
comparison group; they were selected as R1-negative under Q1, so they are not a calibration set
with a known null. Eight of the 12 cells keep R1 under both variants; the three AR-block/DeiT/C-10
cells and the AR-full/DeiT/C-100 MLP cell at $O_2$ each fall from two positive seeds to one, so for
them R1 depends on the working model. None of the 24 comparison cells meets R1 under any of the
three models, although 8 to 10 of their 72 seed tests are flagged. This shows stability of the
eight cells to the two specified variants, with the MLP under its original selection; it does not
test the Q1 mean independently, since Q3 retains it.

A separate sensitivity analysis varies the MLP selection rule under Q1.
\ObsResultsSentence{} Neither these findings nor the synthetic panels determine the power of R1 or
R2 at the effect sizes seen on real labels.

\subsection{The pre-specified screen across 22 families}
\label{sec:apparent}\label{sec:taxonomy}

Under accuracy selection, unadjusted MLP gains reach 0.069 nats, but no family passes the
complete screen at any view under either probe, while the output positive control passes in all
22 families (Fig.~\ref{fig:taxonomy}, Table~\ref{tab:taxonomy}). Six family--probe--view
combinations pass a standalone shuffle contrast without passing the screen. External V-MoE
checkpoints and a raw-routing-weight panel keep their original analyses
(Apps.~\ref{app:vmoe},~\ref{app:rawalpha}); the conditional test was not run on them.

\section{What a routing-information claim should report}
\label{sec:standard}

A probe's practical benefit and an interpretation of its features are separate claims, and a fair
held-out comparison establishes only the first. For the second, our results suggest making four
choices explicit; they are reporting recommendations, not conditions that guarantee a correct
conclusion.

\textbf{(1) Fitting and selection.} Report the training loss, validation criterion and checkpoint
rule of both probes, and choose one proper selection score in advance while reporting both
evaluation scores. Selecting on a proper score is standard advice; what the benchmark adds is its
cost when violated, 50/120 against 0/120 detections on fixed trajectories under a known null. Any
comparison in which two feature sets follow different training trajectories can be affected, but
this has been measured here for two MLP implementations, not in general.
\textbf{(2) The reference.} State the reference distribution and its assumptions. Shuffling routing
globally breaks its dependence on the outputs and answers a different question. For a learned
statistic such as ours, the references must go through the same fitting and selection procedure
as the observed assignment.
\textbf{(3) The representations.} State what the output view and the routing summary contain;
sorted probabilities and logits discard class identity, and entropy profiles discard trace detail.
\textbf{(4) Operating characteristics of the rule actually used}, over an effect-size range
relevant to the question, stating how the benchmark effect relates to the observed statistic. The
sensitivity of the complete decision rule must be measured directly. Single-test power does not
determine the power of a rule combining two scores and requiring at least two positive training
seeds. Table~\ref{tab:selfcheck} records these choices for this audit.

\section{Related work}
\label{sec:related}

\paragraph{Uncertainty from outputs and internal decisions.}
Confidence-based error prediction and calibration motivate the output baselines
\citep{hendrycks2017baseline,guo2017calibration}, and proper scores evaluate probabilities without
binning choices \citep{gneiting2007proper}. Attention-residual aggregation \citep{kimi2026attnres},
expert routing \citep{shazeer2017moe,fedus2022switch,riquelme2021vmoe}, adaptive computation
\citep{graves2016act} and early exits \citep{schuster2022calm} supply the internal features. Gate
entropy has been used for segmentation uncertainty \citep{pavlitska2025extracting} and
out-of-distribution detection \citep{lee2026unseen}; such work targets useful prediction, which
does not require the conditional-information claim examined here.

\paragraph{Conditional probing and prior routing audits.}
Usable information depends explicitly on the predictor class \citep{xu2020usable}.
\citet{hewitt2021conditional} define conditional probing as usable (V-)information beyond a
baseline, estimated by comparing concatenated features with a padded baseline, the same
construction used here; our population example is consistent with that framing, in which a
restricted-risk gain need not be a Shannon-information gain. Control tasks
\citep{hewitt2019control}, MDL probes \citep{voita2020mdl} and information-theoretic probing
\citep{pimentel2020information} address related interpretation questions.
\citet{liang2026routing} already identify capacity and assignment confounds in attention-residual
routing probes. The additions here are the measured behaviour of raw and shuffle comparisons under
a label null that holds by construction, a fixed-trajectory intervention on checkpoint selection in
two implementations, and the sensitivity of the corrected comparison relative to a conditional
reference.

\paragraph{Conditional references and predictive importance.}
The conditional permutation test \citep{berrett2020conditional} and model-X inference
\citep{candes2018knockoffs} rely on a known or well-estimated conditional law of the tested
feature; we reuse the former rather than propose a new test. The hardness of unrestricted
conditional-independence testing \citep{shah2020hardness} motivates reporting diagnostics, null
behaviour and sensitivity. \citet{mentch2022getting} show that pure noise can improve out-of-sample
bagging predictions and register as important, and \citet{williamson2023importance} define
algorithm-agnostic importance through oracle predictive contrasts. Our measurements concern
routing-probe comparisons under natural output--routing dependence.

\section{Limitations}
\label{sec:limitations}

The scope is in-distribution top-1 correctness in image classification. Sorted outputs omit class
identity and routing entropies omit trace detail. All R1-positive families are DeiT
attention-residual, and coverage of MoE, halting and external pretrained models is limited;
language-model routing, distribution shift and epistemic--aleatoric decompositions are not
studied. All 66 recoverable training configurations include a soft-binned calibration penalty and
two are unknown, so the results are not an audit of arbitrary cross-entropy training, and the
pooled runs are not a controlled comparison of architecture quality (App.~\ref{app:config}).

The label null uses one family of label laws, a logistic function of $O_0$. At that view the
output-only logistic class contains the generator, so the benchmark does not exercise the
approximation route of Proposition~\ref{prop:gap}, and it cannot certify the conditional model for
other dependences between correctness and outputs. The selection mechanism was isolated at $O_0$
on six models, in two MLP implementations. The log-loss policy is applied to every
common-generator MLP evaluation and all 68 observed-label MLP runs; the pre-specified screen,
the raw-weight and external panels and the independent-split power curves keep the original
accuracy-based rule. Linear probes involve no validation-based selection.

The conditional test is exact only under the true conditional law. R1 in four of the 12 cells,
including every AR-block/DeiT/C-10 cell, does not persist under the two working-model variants
(App.~\ref{app:qsens}). The doubled-$S$ check fails its one-step criterion on the checked cell and
changes one test decision there, so individual conditional-test decisions can depend on the
sampler length; one paired-row discriminator (D3) is excluded because its cross-validation let
paired counterparts cross folds (App.~\ref{app:cptdiagnostics}). Neither diagnostic establishes
adequacy of $\hat q$. The common-generator
and historical fold-specific benchmarks share permutation draws, so they form a paired
re-analysis rather than independent replications (App.~\ref{app:seeds}). Primary tests use 5,000
examples, bootstrap intervals condition on fitted predictions, runs share evaluation images, and
R1/R2 carry no multiplicity guarantee. Non-detection cannot establish output sufficiency.

\section{Conclusion}
\label{sec:conclusion}

A routing feature can improve a fitted probe without adding information beyond the chosen outputs.
Under a label null that preserves real output--routing dependence, the width-matched MLP comparison
reported such gains in half of the confidence-only evaluations, and holding training trajectories
fixed traced them to accuracy-based checkpoint selection in two independent implementations. Aligning
selection with the reported score removed these detections but did not make the comparison
sensitive to small signals in the tested settings. A conditional reference asks a different
question, and what it can show depends on the conditional model and on the measured sensitivity of
the complete decision rule.

On real correctness labels, the primary analysis finds conditional-assignment evidence in five
DeiT attention-residual families, relative to an estimated conditional law. In four of these
families the evidence persists under two specified variants of that law, and no family passes an
additional noise criterion. These findings establish neither general routing information nor output
sufficiency. Together, these results show that evidence for information beyond model outputs has to account for
probe fitting, the reference distribution and the behaviour of the whole decision rule, not only
for the number of probe parameters.

\phantomsection\label{sec:stmts}
\subsection*{AI use statement}
Generative AI tools were used throughout this project: to propose and refine hypotheses and
analysis designs, to give feedback on experimental methodology, to implement, run and monitor
experiment and analysis code, to interpret results, to search and summarise related literature,
to prepare figures, to draft and edit the manuscript, and to check its internal consistency. This
included designing, specifying before execution, implementing and running the checkpoint-selection
experiments of \S\ref{sec:mechanism} (including the independent PyTorch implementation), the
benchmark-wide log-loss re-analysis, the working-model sensitivity analysis of
\S\ref{sec:cptresults}, the doubled-$S$ check and the seed audit of App.~\ref{app:seeds}. Every
reported experimental value is computed from stored experiment records by the scripts in the
supplementary material, and post-hoc analyses are labelled as such. The authors are responsible
for the protocols, code, numerical results, interpretations and manuscript.

\subsection*{Ethics statement}
The study uses image-classification benchmarks and involves no human-subject experiment or
deployment. Conclusions concern the evaluated representations, models and estimators;
non-detection should not be used to dismiss routing signals outside this scope.

\subsection*{Reproducibility statement}
A code-and-records archive accompanying this paper, available from the authors, contains the
paper source, all analysis code, the protocols
(including those of the post-hoc analyses), execution logs, integrity manifests, aggregate results
for every table, the per-example caches of the six benchmark models, and the per-replicate records
of the label-null benchmark, the checkpoint-selection experiments, the log-loss re-analysis, the
independent-split benchmarks and the conditional tests on real labels. Its README gives tested
commands for integrity checks, a recount of the headline numbers from these records, a
recomputation of the conditional-test $p$-values of the eight primary and revision record types listed
in the README (9,932 two-score tests) from their stored reference statistics, unit tests,
regeneration of the tables of this revision, Figure~\ref{fig:mech} and this PDF, and a minimal
end-to-end refit of one benchmark replicate. It also lists what a complete rerun would need in
addition (the caches of the other 62 runs, some per-replicate records, checkpoints and images),
which are omitted for size and identified by SHA-256.
App.~\ref{app:protocol} specifies the screening rule; Apps.~\ref{app:metric}, \ref{app:t4} and
\ref{app:cpt} describe the checkpoint intervention, the exact label-null benchmark and the
conditional analysis; App.~\ref{app:provenance} records provenance, corrections and exclusions.
The proof of Proposition~\ref{prop:gap} is in App.~\ref{app:proof}.

\phantomsection\label{sec:refs}
\bibliography{refs}
\bibliographystyle{iclr2027_conference}
\appendix
\setlength{\emergencystretch}{2em}
\phantomsection\label{sec:appstart}
\section{Pre-specified protocol and taxonomy rules}
\label{app:protocol}

The primary decision rule and taxonomy below were specified before the confirmatory runs were
analysed and were not subsequently modified. Additional analysis amendments that did not change
the primary decision rule are recorded in Table~\ref{tab:auditlog}.

\textbf{Estimand.} $T=\mathbf{1}\{\text{argmax correct}\}$; $O_j$ an output view;
$R$ the routing trace. Our raw probe statistic for $\mathrm{Gain}_S(R\mid O_j)$ is the paired
held-out improvement of a probe on $(O_j, R)$ over the same probe on $O_j$ zero-padded to
identical input dimension, for $S\in\{$log loss, Brier$\}$. It is a finite-probe statistic, not
the Bayes-risk quantity itself.

\textbf{Output views.} $O_0$ top-class confidence (as a logit); $O_1$ adds predictive entropy,
top-two probability margin, logit $L_2$ norm and runner-up probability; $O_2$ the sorted
probability vector; $O_3$ the sorted logit vector. Sorting makes the view permutation-invariant;
class-identity features are a logged exclusion.

\textbf{Probes.} Linear: logistic regression, $C=1.0$, $\le 3000$ iterations. Nonlinear: one
hidden layer of 64 units, $\alpha=10^{-3}$, early stopping with patience 15, $\le 300$
iterations, fixed seed. Hyperparameters are identical for every feature set, so search budget
cannot differ between routing and baseline. Features are standardised using train-fold
statistics only.

\textbf{Cross-fitting.} Five-fold stratified splits $\times$ two repeats (fold seeds 100, 101),
identical folds for every feature set; out-of-fold predictions averaged over repeats.

\textbf{Controls.} (i) architecture- and dimension-matched zero-padded $O_j$; (ii) real $R$;
(iii) shuffled $R$, permuted independently within train and evaluation folds, preserving all
marginals and the joint geometry while destroying the sample-level correspondence;
(iv) $\varepsilon\sim\mathcal{N}(0,I)$ of matching dimension.

\textbf{Detection.} A comparison is detected when the 95\% percentile bootstrap interval over
2000 example resamples excludes zero positively in at least two training seeds for each
proper score. The stored original aggregation counts passing seeds separately for log loss
and Brier, so the passing seed sets need not coincide. The R2 rule explicitly requires
the two CPT scores and noise contrast to pass within the same run before seed aggregation. Example-level and seed-level uncertainty are kept separate;
examples are never pooled across seeds.

\textbf{Taxonomy.} A family is \textbf{Type\,I} if the raw gain, the real-vs-shuffle contrast
and the real-vs-noise contrast are all detected at view $O_0$ and no complete three-way pass occurs at a higher designated view;
\textbf{Type\,II} if the same triple is detected at $O_1$ but neither $O_2$ nor $O_3$; \textbf{Type\,III} if at $O_2$ or
$O_3$; \textbf{Type\,0} otherwise. A family whose positive control is undetected cannot receive
a Type\,0 verdict.

\Needspace{330pt}
\section{Self-assessment against the reporting standard}
\label{app:self-checklist}

\begin{table}[h]
\centering\footnotesize
\setlength{\tabcolsep}{4pt}
\caption{The audit against the four requirements of \S\ref{sec:standard}: what was done for each, and the inferential limit that remains.}
\label{tab:selfcheck}
\begin{tabular}{p{0.30\linewidth}p{0.64\linewidth}}
\toprule
\textbf{Requirement (\S\ref{sec:standard})} & \textbf{This study} \\
\midrule
Fit and select both probes by the score you report & Log loss and Brier co-primary, out-of-fold (5-fold $\times$ 2 repeats); probes dimension-matched by zero-padding, which does not match fitting (Hazard~I, App.~\ref{app:hazards}). The original MLP selected checkpoints by validation accuracy; the corrected log-loss policy and its benchmark-wide re-analysis are in App.~\ref{app:metric}. \\
Use a conditional reference & Estimated-$q$ CPT with 79 references (App.~\ref{app:cpt}); GLOBAL and post-hoc MATCHED shuffles are assignment controls and the Gaussian block a representation control (Hazard~II); standalone shuffle detections in five families, none meeting the frozen screen. \\
State what the output view contains & Four views $O_0$--$O_3$; sorted views omit class identity; routing representation stated (entropy profile; raw $\alpha$ on 12 runs, App.~\ref{app:alpha}). \\
Report operating characteristics for the rule you actually use & Common-generator null rates (App.~\ref{app:t4}), independent-split power at $O_1$ and the matched panel (App.~\ref{app:cpt}, \ref{app:metric}); three seeds per family (five for one); output positive control 22/22; family-rule power of R1/R2 not established. \\
\bottomrule
\end{tabular}
\vspace{2pt}\begin{minipage}{\linewidth}\footnotesize
\textit{Note.} A reported component check does not establish the power or validity of the full
replicated rule.\end{minipage}
\end{table}

\FloatBarrier
\section{Per-family results at every view}
\label{app:perfamily}
\begin{table}[!htbp]
\centering
\footnotesize\renewcommand{\arraystretch}{0.95}
\setlength{\tabcolsep}{3.2pt}
\caption{\textbf{Positive controls pass, but no family meets the frozen full criterion.} Panel A summarizes 22 families and 68 training runs; Panel B separately reports two public pretrained checkpoints. Gains are held-out log-loss improvements in nats.}
\label{tab:taxonomy}
\begin{tabularx}{\linewidth}{@{}Xrrrrr@{}}
\toprule
\textbf{Mechanism} & \textbf{Families} & \textbf{Runs} & \shortstack{Positive-control\\gain} & \shortstack{Max naive\\gain} & \shortstack{Frozen full\\pass} \\
 & & & \scriptsize(range) & \scriptsize(max) & \scriptsize(families) \\
\midrule
Attention-residual\newline block   & 9 & 27 & $+0.0037$ to $+0.0149$ & $+0.0561$ & 0 / 9 \\
Attention-residual\newline full    & 9 & 29 & $+0.0037$ to $+0.0189$ & $+0.0337$ & 0 / 9 \\
MoE top-1 gate              & 2 &  6 & $+0.0063$ to $+0.0107$ & $+0.0693$ & 0 / 2 \\
Adaptive-depth\newline halting      & 2 &  6 & $+0.0088$ to $+0.0088$ & $+0.0511$ & 0 / 2 \\
\midrule
\textbf{Total} & \textbf{22} & \textbf{68}
 & \textbf{22/22 detected} & \textbf{up to $+0.069$} & \textbf{0 / 22} \\
\bottomrule
\end{tabularx}

\vspace{2pt}
\begin{tabularx}{\linewidth}{@{}Xrrrrr@{}}
\multicolumn{6}{l}{\textbf{B. External pretrained validation} (not members of the frozen taxonomy)} \\
\toprule
\textbf{Checkpoint} & \textbf{top-1} & \textbf{MoE layers}
 & \shortstack{Positive-control\\gain} & \shortstack{Standalone\\hits} & \shortstack{Full\\pass} \\
 & \scriptsize(50k ImageNet) & & \scriptsize(linear / MLP) & \scriptsize(of 8) & \scriptsize(of 8) \\
\midrule
V-MoE B/16 & $0.8554$ & 6 & $+0.0006$ / $+0.0007$ & 2 & 0 / 8 \\
V-MoE S/32 & $0.7578$ & 2 & $+0.0018$ / $+0.0007$ & 1 & 0 / 8 \\
\bottomrule
\end{tabularx}
\vspace{2pt}
\begin{minipage}{\linewidth}\footnotesize
\textit{Note.} A family is a routing-variant/backbone/dataset cell, with three training
seeds (five for DeiT/C100/AR-full). Positive-control gain is Linear $O_1$ versus $O_0$;
max naive gain is the largest RAW MLP gain at $O_0/O_1$. The full criterion requires RAW,
GLOBAL and noise contrasts at the same view, both scores and seed replication.
Type 0 means this criterion was not met, not zero incremental information or output
sufficiency: all 22 families are Type 0, with no Type I--III assignments.
The independent-split linear CPT endpoint is in \S\ref{sec:power}.
Panel B checkpoints are not members of the 22/68 panel and have no multi-seed requirement.
\end{minipage}
\end{table}

\begin{figure}[htbp]
\centering\includegraphics[width=\linewidth]{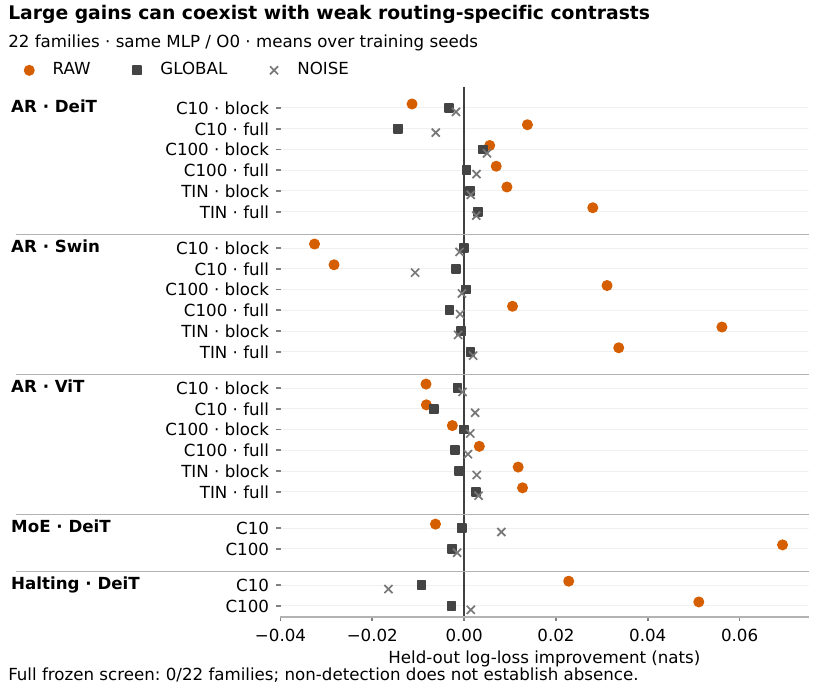}
\caption{\CapFigTaxonomy}\label{fig:taxonomy}
\end{figure}
\begin{table}[htbp]\centering\small\setlength{\tabcolsep}{4pt}
\caption{Independent-split operating characteristics at view $O_1$: $\hat q$ is fitted on 5,000
examples and every statistic is evaluated on the 5,000 disjoint examples. MLP rows use the
original accuracy-based stopping and selection; linear rows involve no validation-based
selection. Columns are the reference gain of the \emph{synthetic} routing signal, a model-based
quantity that is not an observed conditional-reference contrast. Rows are per-replicate single tests, not the cross-seed R1 or
R2 family rules; the linear rows use 200 replicates and the MLP rows 50. The second block
is the stored conjunction, CPT $\wedge$ noise on the same test rows: a per-replicate quantity,
not the cross-seed R2 rule.}
\label{tab:power}
\begin{tabular}{lllrrrrrr}\toprule
Statistic & Cell & Probe & \multicolumn{6}{c}{Reference gain of the synthetic signal (nats)} \\
\cmidrule(lr){4-9}
 & & & 0 (null) & 0.0005 & 0.001 & 0.002 & 0.005 & 0.01 \\\midrule
CPT & DeiT/C-100/AR-block & linear & 4/200 & 14/200 & 32/200 & 86/200 & 191/200 & 200/200 \\
CPT & DeiT/C-100/AR-block & mlp & 1/50 & 2/50 & 2/50 & 6/50 & 19/50 & 46/50 \\
CPT & Swin/C-10/AR-block & linear & 1/200 & 25/200 & 51/200 & 129/200 & 200/200 & 200/200 \\
CPT & Swin/C-10/AR-block & mlp & 0/50 & 0/50 & 1/50 & 3/50 & 8/50 & 21/50 \\
CPT $\wedge$ noise & DeiT/C-100/AR-block & linear & 1/200 & 1/200 & 4/200 & 22/200 & 133/200 & 198/200 \\
CPT $\wedge$ noise & DeiT/C-100/AR-block & mlp & 0/50 & 1/50 & 1/50 & 2/50 & 11/50 & 37/50 \\
CPT $\wedge$ noise & Swin/C-10/AR-block & linear & 0/200 & 8/200 & 15/200 & 44/200 & 162/200 & 200/200 \\
CPT $\wedge$ noise & Swin/C-10/AR-block & mlp & 0/50 & 0/50 & 1/50 & 3/50 & 8/50 & 20/50\\
\bottomrule\end{tabular}\end{table}

Table~\ref{tab:perfamily} gives, for every family, the output positive-control gain and the more
favourable GLOBAL contrast of the two probes at each view.

\begin{table}[!ht]
\centering\footnotesize
\setlength{\tabcolsep}{3.4pt}
\caption{No family meets the original full criterion despite positive controls in every row. All 22 families are retained, with held-out log-loss improvements in nats.}
\label{tab:perfamily}
\begin{tabular}{lllrrrrrr}
\toprule
\textbf{Mechanism} & \textbf{Backbone} & \textbf{Data} & \textbf{Runs}
 & \textbf{Pos.\ ctrl} & $O_0$ & $O_1$ & $O_2$ & $O_3$ \\
\midrule
AR-block & Swin-T & C-10 & 3 & $+0.0037$ & $+0.0004$ & $+0.0161$ & $-0.0003$ & $-0.0000$ \\
AR-full & Swin-T & C-10 & 3 & $+0.0037$ & $+0.0010$ & $+0.0152$ & $+0.0048$ & $-0.0003$ \\
AR-block & Swin-T & C-100 & 3 & $+0.0075$ & $+0.0008$ & $+0.0001$ & $+0.0006$ & $+0.0019$ \\
AR-full & Swin-T & C-100 & 3 & $+0.0100$ & $+0.0006$ & $-0.0004$ & $-0.0001$ & $-0.0007$ \\
AR-block & DeiT-S & C-10 & 3 & $+0.0050$ & $+0.0008$ & $-0.0003$ & $+0.0092$ & $-0.0000$ \\
AR-full & DeiT-S & C-10 & 3 & $+0.0064$ & $+0.0007$ & $+0.0000$ & $-0.0002$ & $-0.0001$ \\
AR-block & DeiT-S & C-100 & 3 & $+0.0108$ & $+0.0040$ & $+0.0005$ & $+0.0006$ & $+0.0006$ \\
AR-full & DeiT-S & C-100 & 5 & $+0.0127$ & $+0.0012$ & $+0.0003$ & $+0.0001$ & $+0.0004$ \\
AR-block & Swin-T & T-IN & 3 & $+0.0083$ & $-0.0002$ & $+0.0012$ & $-0.0003$ & $+0.0001$ \\
AR-full & Swin-T & T-IN & 3 & $+0.0135$ & $+0.0016$ & $-0.0004$ & $-0.0003$ & $-0.0003$ \\
AR-block & DeiT-S & T-IN & 3 & $+0.0149$ & $+0.0013$ & $-0.0003$ & $-0.0002$ & $-0.0006$ \\
AR-full & DeiT-S & T-IN & 3 & $+0.0189$ & $+0.0030$ & $+0.0017$ & $+0.0008$ & $+0.0006$ \\
AR-block & ViT-B & C-10 & 3 & $+0.0065$ & $+0.0007$ & $-0.0003$ & $-0.0008$ & $-0.0004$ \\
AR-full & ViT-B & C-10 & 3 & $+0.0064$ & $+0.0013$ & $+0.0001$ & $-0.0004$ & $+0.0135$ \\
AR-block & ViT-B & C-100 & 3 & $+0.0059$ & $+0.0011$ & $+0.0002$ & $-0.0003$ & $+0.0000$ \\
AR-full & ViT-B & C-100 & 3 & $+0.0071$ & $-0.0000$ & $-0.0005$ & $-0.0007$ & $-0.0011$ \\
AR-block & ViT-B & T-IN & 3 & $+0.0084$ & $+0.0003$ & $+0.0001$ & $+0.0000$ & $-0.0003$ \\
AR-full & ViT-B & T-IN & 3 & $+0.0094$ & $+0.0025$ & $-0.0003$ & $-0.0007$ & $-0.0005$ \\
MoE gate & DeiT-S & C-10 & 3 & $+0.0063$ & $-0.0001$ & $+0.0061$ & $-0.0002$ & $+0.0043$ \\
MoE gate & DeiT-S & C-100 & 3 & $+0.0107$ & $+0.0005$ & $+0.0005$ & $+0.0002$ & $+0.0003$ \\
Halting & DeiT-S & C-10 & 3 & $+0.0088$ & $-0.0009$ & $-0.0010$ & $-0.0009$ & $+0.0008$ \\
Halting & DeiT-S & C-100 & 3 & $+0.0088$ & $-0.0005$ & $-0.0002$ & $-0.0005$ & $-0.0004$ \\
\bottomrule
\end{tabular}
\vspace{2pt}
\begin{minipage}{\linewidth}\footnotesize
\textit{Note.} Positive control is $\mathrm{Gain}(O_1\mid O_0)$. Other columns show the more favourable GLOBAL contrast of the two probes at each view; this descriptive selection is not the full criterion. Leaving Type 0 requires RAW, GLOBAL and noise detections at the same view, both scores and seed replication.
\end{minipage}
\end{table}

\section{Raw routing-weight panel}
\label{app:alpha}\label{app:rawalpha}

For a matched panel of twelve runs in four families (AR-block/DeiT on C-10 and C-100,
AR-full/DeiT/C-100 and AR-block/Swin/C-100) we extracted the token-averaged routing distribution
of every attention-residual layer and concatenated it, giving $D_\alpha=154$ (both DeiT AR-block
families), $299$ (AR-full/DeiT) and $20$ (AR-block/Swin) dimensions, against entropy profiles of
$22$, $23$ and $8$ dimensions. Extractions were validated against the stored evaluation records: accuracy
agreed to four decimals in 12/12 runs, and per-layer entropies recomputed from $\alpha$ ranked
with the stored profile at Spearman $0.977$--$0.991$.

Running the identical set of output views on $R_\alpha$: the head-to-head comparison against the entropy
profile at equal padding is $\le 0$ in all four families and both probe classes (for example
$-0.0046$ linear and $-0.0127$ nonlinear on AR-block/DeiT/C-100), and no family attains a
Type\,II or Type\,III assignment with the richer trace. Under the pre-specified probe class and
available sample sizes, access to raw routing weights does not improve these fitted probes over
the entropy profile. We do not read this as evidence that the compression is
information-preserving: the higher-dimensional trace is also harder to estimate at fixed $n$,
and separating the two explanations would require a dimension-controlled study we did not run.

This panel also supplies the sharpest form of Hazard~II: with $R_\alpha$, real routing beats the
Gaussian control at 3/3 seeds in three of the four families ($+0.0169$, $+0.0292$, $+0.0164$
nats) while beating shuffled routing in none of them.

\section{Proof of Proposition~\ref{prop:gap}}
\label{app:proof}

\textbf{Construction.} Let $O$ be uniform on $\{-1,0,+1\}$, let $\phi(o)=o^2$ and $R=\phi(O)$, and
let $T\mid O=o \sim \mathrm{Bern}(\eta(o))$ with $\eta(o)=\sigma(o^2)$, so
$\eta(\pm1)=\sigma(1)\approx0.7311$ and $\eta(0)=\tfrac12$. All probabilities lie in $(0,1)$ and
$O$ has three support points; both facts are used below.

\textbf{Step 1: $T\perp R\mid O$ and the population gain is zero.} Conditionally on $O=o$ the
variable $R$ equals the constant $\phi(o)$, and every random variable is independent of a
degenerate one, so $P(T\in\cdot,R\in\cdot\mid O)=P(T\in\cdot\mid O)\,P(R\in\cdot\mid O)$. Hence
$T\perp R\mid O$. Moreover $I(T;R\mid O)\le H(R\mid O)=0$, and $\eta_{OR}(o,\phi(o))=\eta_O(o)$
almost surely, so by strict propriety $\mathrm{Gain}_S(R\mid O)=0$ for \emph{every} strictly
proper $S$, not only for log loss.

\textbf{Step 2: the augmented class attains the Bayes risk.} Taking $(a,c,b)=(0,1,0)$ gives
$q(o,r)=\sigma(r)=\sigma(o^2)=\eta(o)$. Hence
$\inf_{\mathcal{F}_{O,R}}\mathcal{R}_{\log}=\mathcal{R}^\star_{\log}=\mathbb{E}[H(\eta(O))]$, and
the infimum is attained.

\textbf{Step 3: exact excess risk of the baseline class.} For $q_{a,b}=\sigma(aO+b)$ the pointwise
decomposition of log loss into entropy plus divergence gives
\[
\mathcal{R}_{\log}(q_{a,b})-\mathcal{R}^\star_{\log}
=\Delta(a,b):=\tfrac13\!\!\sum_{o\in\{-1,0,1\}}\!\!\mathrm{KL}\big(\eta(o)\,\big\|\,\sigma(ao+b)\big).
\]
Write $p=\sigma(1)$. Then
$3\Delta(a,b)=\mathrm{KL}(p\|\sigma(b-a))+\mathrm{KL}(\tfrac12\|\sigma(b))+\mathrm{KL}(p\|\sigma(b+a))$.
The map $z\mapsto\mathrm{KL}(p\|\sigma(z))=\text{const}+p\log(1+e^{-z})+(1-p)\log(1+e^{z})$ is
convex, so $f(a)=\mathrm{KL}(p\|\sigma(b-a))+\mathrm{KL}(p\|\sigma(b+a))$ is convex; it is also
even. A convex even function on $\mathbb{R}$ satisfies $f(0)\le\tfrac12(f(a)+f(-a))=f(a)$, so
$f$ is minimised at $a=0$. The middle term does not depend on $a$. Hence
$\Delta(a,b)\ge\Delta(0,b)$ for all $a$, and the best $O$-only logistic probe is a
\emph{constant} predictor $t=\sigma(b)$.

\textbf{Step 4: the best constant and the value of the gap.} Minimising
$t\mapsto\sum_o w_o\,\mathrm{KL}(\eta(o)\|t)$ over $t\in(0,1)$ gives first-order condition
$\sum_o w_o\eta(o)=t$, so $t^\star=\mathbb{E}[\eta(O)]=\tfrac13(2\sigma(1)+\tfrac12)\approx0.6540$,
which is interior and therefore attained at $b=\operatorname{logit}t^\star$. Using
$\sum_o w_o\mathrm{KL}(\eta(o)\|t^\star)=H(t^\star)-\sum_o w_o H(\eta(o))$,
\[
\inf_{\mathcal{F}_O}\mathcal{R}_{\log}-\inf_{\mathcal{F}_{O,R}}\mathcal{R}_{\log}
= H(t^\star)-\mathbb{E}\!\left[H(\eta(O))\right] = I(T;O)\approx 0.0257\ \text{nats},
\]
strictly positive because $\eta$ is non-constant. \hfill$\square$

\begin{remark}[Beyond log loss]
The strict separation extends to continuous strictly proper scoring rules for which the induced
risk is continuous on the relevant compact prediction set. For such an $S$, let $K$ be the closure of
$\{(\sigma(b-a),\sigma(b),\sigma(b+a)):(a,b)\in\mathbb{R}^2\}$ in $[0,1]^3$. For every $(a,b)$, the middle coordinate lies between the first and third because
$\sigma$ is monotone. This property is closed under limits, so it holds throughout $K$.
The target $\eta=(p,\tfrac12,p)$ with $p=\sigma(1)>\tfrac12$ violates it. Hence
$\eta\notin K$. $K$ is compact and the risk continuous on it, so the
infimum over $K$ is attained at some $\bar q\neq\eta$, and strict propriety makes it exceed the
Bayes risk. For the Brier score the gap is $0.0119$. We do not claim this for every strictly
proper score: the argument uses continuity of the risk on $K$, which excludes rules whose risk is
unbounded at the boundary. The zero-gain half of Proposition~\ref{prop:gap} needs no such
condition and holds for every strictly proper $S$, since $\eta_{OR}=\eta_O$ almost surely.
\end{remark}

\begin{remark}[The redundancy need not be exact]
Let $R_\delta=\phi(O)+\delta Z$ with $Z$ uniform on $[-1,1]$ independent of $(T,O)$. The law of
$R_\delta$ given $O$ does not depend on $T$, so $T\perp R_\delta\mid O$ remains exact and
$\mathrm{Gain}_S=0$. All scores are bounded on the relevant compact range, so
$\mathcal{R}_{\log}$ at $(a,c,b)=(0,1,0)$ is continuous in $\delta$ and converges to
$\mathcal{R}^\star_{\log}$ as $\delta\to0$; hence the strict gap persists for all sufficiently
small $\delta$. It degrades smoothly rather than vanishing: the gap is $0.0257$, $0.0237$,
$0.0190$, $0.0105$ and $0.0038$ nats at $\delta=0,\,0.25,\,0.5,\,1,\,2$ (400-node Gauss--Legendre
quadrature in $Z$; $\delta=0$ agrees with the closed form to $10^{-12}$). Deterministic redundancy
is the cleanest instance of the phenomenon, not a necessary condition for it.
\end{remark}

\begin{remark}[Scope]
The gap is a property of the probe class. If $\mathcal{F}_O$ were closed under substituting
$R=\phi(O)$ --- for instance the class of all measurable functions of $O$ --- both infima would
equal the Bayes risk and the gap would be $0$. The proposition therefore does not say that probes
are uninformative; it says that a probe improvement is not a functional of $(T,O,R)$ alone, so it
cannot by itself certify $\mathrm{Gain}_S(R\mid O)>0$. It is a population approximation statement
and is distinct from the finite-sample behaviour measured in \S\ref{sec:exactnull}, which uses a
far richer probe class and no exact redundancy.
\end{remark}

\Needspace{6\baselineskip}
\section{Mechanism diagnostics for the two hazards}
\label{app:hazards}

These decompositions examine the probe behaviour of \S\ref{sec:exactnull} and are retained
as mechanism diagnostics rather than as headline evidence.

\begin{figure}[t]
\centering\includegraphics[width=\linewidth]{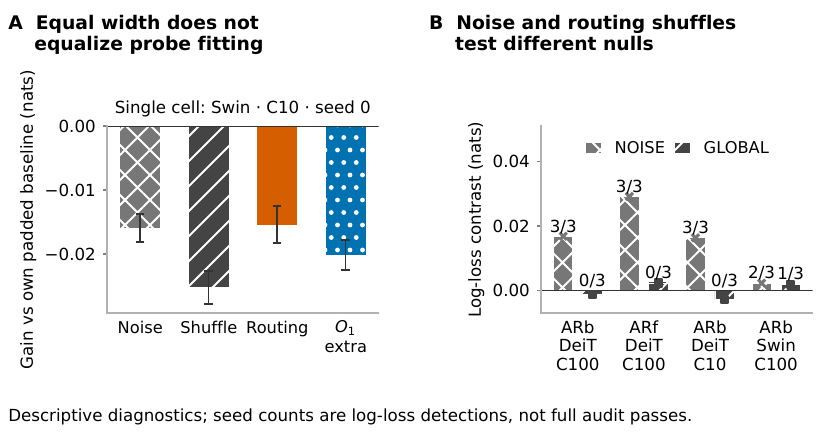}
\caption{\CapFigHazards}\label{fig:hazards}
\end{figure}

Figure~\ref{fig:hazards} summarises these comparisons.

\textbf{Hazard~I: capacity matching is not optimisation matching.} Padding equalises parameter
count and architecture but not training dynamics. In a one-cell diagnostic under the original
accuracy-selected MLP, appending \emph{any} real-valued block --- Gaussian noise, shuffled routing, real
routing, or genuinely informative output features --- degrades held-out score relative to the zero-padded twin
by comparable amounts ($-0.016$, $-0.025$, $-0.015$, $-0.020$ nats). At sweep level, $6$ of the
$22$ families have a \emph{negative} best naive MLP gain, corroborating that the padded-baseline
statistic can change sign across feature representations. This pattern may be related to the
selection effect of \S\ref{sec:mechanism}, which was established later and on a different panel;
it was not separately isolated here. We do not read the single-cell
decomposition as establishing a universally replicated mechanism.

\textbf{Hazard~II: a noise control is not a routing-assignment control.} A Gaussian random-feature
control asks whether extra unstructured dimensions help. A shuffle preserves the empirical
distribution and internal geometry of $R$ while destroying its original sample assignment; unlike
Gaussian noise it therefore controls for routing-like feature structure, although the global
shuffle also disrupts the dependence between $R$ and $O$ and is not an exact conditional null
(\S\ref{sec:ria}). The two disagree in practice: real routing beats Gaussian noise at 3/3 seeds in
three of four matched-panel families while beating shuffled routing in none, and in the full sweep
two families (AR-full/DeiT/T-IN, AR-block/Swin/T-IN) pass the noise contrast and fail the shuffle
contrast. A noise-only screen would flag those families despite the lack of assignment evidence.

\Needspace{8\baselineskip}
\section{Historical fold-specific label benchmark and its qualification}
\label{app:exactnull}

This historical benchmark was specified before its outcomes and is retained for transparency.
It used fold-specific output-only generators, so it does not by itself establish
$T^*\perp R\mid O_j$. The common-generator benchmark was pre-specified to resolve this
implementation-level exactness issue and supplies the primary null-validity evidence
(App.~\ref{app:t4}). The historical and common-generator comparisons agree qualitatively on nonlinear
comparator inflation and lower CPT detection rates; local control ordering is not invariant.
No historical count or taxonomy result is changed. Table~\ref{tab:nullbench} summarises the
three label-null benchmarks used in this paper.

\begin{table}[htbp]\centering\footnotesize\setlength{\tabcolsep}{3pt}
\caption{The three label-null benchmarks. Only the common-generator benchmark draws labels from
a single frozen function of $O_0$, which is what makes its null exact conditional on that generator.}
\label{tab:nullbench}
\begin{tabular}{>{\raggedright\arraybackslash}p{0.19\linewidth}>{\raggedright\arraybackslash}p{0.25\linewidth}>{\raggedright\arraybackslash}p{0.17\linewidth}>{\raggedright\arraybackslash}p{0.10\linewidth}>{\raggedright\arraybackslash}p{0.19\linewidth}}\toprule
Benchmark & Label generator & Evaluation size & Exact null? & Primary role \\\midrule
Historical full-sample (App.~\ref{app:exactnull}) & fold-specific $\hat\eta_{F_i}(O_{0,i})$, seed-11 folds & 10,000 examples; 6 cells $\times$ 200 draws & no (conditional on fold) & historical null counts, Table~\ref{tab:historicalfull} \\
Historical fold-specific 5k (App.~\ref{app:histcal}) & same recipe on the 5k test half & 5,000 examples; 1,920 label records & no & historical CPT calibration, Tables~\ref{tab:t2nullcontrols}--\ref{tab:t2nullcells} \\
Common-generator (App.~\ref{app:t4}) & one frozen $g(O_0)$ fitted on half A & 5,000 examples; 600 label vectors, 3,840 evaluations & yes, given $g$ & primary null validity; Table~\ref{tab:nllcommon} \\
\bottomrule\end{tabular}\end{table}

\subsection{Original full-10k benchmark}

The design was specified \textbf{before} any replicate was produced or inspected
(App.~\ref{app:provenance}).

\textbf{Cells (6, pre-specified, no post-hoc substitution).} All use training seed 0:
AR-block/DeiT/C100, AR-block/Swin/C10, AR-full/DeiT/TIN, AR-block/ViT/C100,
MoE/DeiT/C10 and halting/DeiT/C100---three backbones, three datasets and all three
mechanism classes. Exact artifact identifiers are in App.~\ref{app:provenance}.

\textbf{Implemented generation rule and qualification.}
Every real pair $(O_i,R_i)$ is retained. A five-fold, correctness-stratified split with seed 11
fits logistic output-only predictors, and example $i$ receives
$g_i=\hat\eta_{F_i}(O_{0,i})$ from its held-out generator fold $F_i$.
Synthetic labels are drawn independently as $T_i^\star\sim\mathrm{Bernoulli}(g_i)$.
There are 200 label draws per cell and 1,200 records in the original experiment.

The pre-specified protocol described this as an exact conditional null given $O_j$. That statement
requires a common function of $O_0$, or further assumptions on fold membership; cross-fitting
alone does not provide it. The implemented generator is a function of $(O_0,F)$, so the
construction directly establishes independence from routing only after additionally conditioning
on $F$ and the fitted generators. The audit probes do not condition on this generator fold.
A stored-output check identified the issue concretely in halting/DeiT/C-100: examples 4654
and 6709 (zero-based example indices) have identical logits and different routing profiles, but
belong to different generator folds. The stored records give different generator probabilities for these identical
output views. Thus the advertised exact-null argument cannot
be applied to this implementation as written. We retain the recorded detection counts and
qualify their interpretation; they measure behaviour under this specified recipe rather than
prove rejection rates under an unrestricted conditional null given $O_j$ alone.

The conditional-permutation analysis uses the same fold-specific generator recipe on its
5k test half and inherits this qualification (App.~\ref{app:cpt}). The independent Gaussian-routing null at
zero injection in the power experiments is unaffected: routing is then independent of all
label-generator inputs. Proposition~\ref{prop:gap} also remains an exact population result.

\textbf{Evaluation.} The same probe classes and fitting budget: 5-fold stratified $\times$ 2 repeats on
identical folds, in-fold standardisation, log loss and Brier, 2000-resample paired example
bootstrap, and the same component detection rule (both proper scores, within a single replicate).
Unlike the original within-fold audit control, this benchmark constructs its routing
permutations once on the full evaluation array before probe cross-fitting. The probability
clipping threshold is $10^{-12}$ rather than the original audit's $10^{-6}$.
Feature sets at identical input dimension: $O_j$ zero-padded, $O_j+R$, $O_j+$ global assignment
shuffle, $O_j+$ output-matched shuffle, $O_j+$ Gaussian block.

\textbf{Storage.} The per-replicate component indicators $D_\mathrm{raw}$, $D_\mathrm{shuf}$,
$D_\mathrm{matched}$, $D_\mathrm{noise}$ are stored individually, so both joint rates are
identifiable afterwards. This is a deliberate correction of the storage defect that made the
conjunction rate unidentifiable in the injection benchmark (App.~\ref{app:provenance}).

\begin{table}[htbp]\centering\footnotesize
\caption{Historical full-sample detections concentrate in the nonlinear probe. Counts at $O_0$ compare the original fold-specific label construction.}\label{tab:historicalfull}
\begin{tabular}{lrrrr}
\toprule
 & \multicolumn{2}{c}{\textbf{Linear}} & \multicolumn{2}{c}{\textbf{MLP}} \\
\cmidrule(lr){2-3}\cmidrule(lr){4-5}
comparison & count & rate & count & rate \\
\midrule
RAW                 & $0/1200$  & $0.000$ & $448/1200$ & $0.373$ \\
GLOBAL              & $10/1200$ & $0.008$ & $165/1200$ & $0.138$ \\
MATCHED             & $5/1200$  & $0.004$ & $149/1200$ & $0.124$ \\
NOISE               & $3/1200$  & $0.003$ & $477/1200$ & $0.398$ \\
joint (RAW\,$\wedge$\,GLOBAL\,$\wedge$\,NOISE)  & $0/1200$ & $0.000$ & $35/1200$ & $0.029$ \\
joint (RAW\,$\wedge$\,MATCHED\,$\wedge$\,NOISE) & $0/1200$ & $0.000$ & $33/1200$ & $0.028$ \\
\bottomrule
\end{tabular}
\vspace{2pt}\begin{minipage}{\linewidth}\footnotesize
\textit{Note.} Denominators are 1,200 replicate--probe evaluations per comparison and
probe; rates retain the recorded three-decimal precision. This historical generator
is qualified in the surrounding text and does not supply the primary exact-label-null evidence.
\end{minipage}\end{table}

Per-cell MLP rates at $O_0$ (raw / global / matched / noise / joint): MoE-C10
$0.055$/$0.250$/$0.205$/$0.270$/$0.005$; Swin-C10 $0.045$/$0.285$/$0.200$/$0.275$/$0.005$;
DeiT-C100 $0.200$/$0.050$/$0.090$/$0.515$/$0.020$; halting-C100
$0.975$/$0.075$/$0.105$/$0.330$/$0.035$; ViT-C100 $0.095$/$0.030$/$0.050$/$0.315$/$0.000$;
DeiT-TIN $0.870$/$0.135$/$0.095$/$0.680$/$0.110$. Global minus matched, in percentage points:
$+0.42$ ($O_0$ linear), $+1.33$ ($O_0$ MLP), $+0.33$ ($O_1$ linear), $+0.42$ ($O_1$ MLP).
Matching locality (median permuted-pair distance in standardised $O_j$): $O_0$ matched $0.0008$
vs global $0.9609$; $O_1$ matched $0.1686$ vs global $2.6884$.

Rates are quoted against a $2.5\%$ \emph{one-sided reference rate}. The joint level of the
two-proper-score rule is not derived, and no nominal level is claimed for it.

\section{External pretrained V-MoE validation}
\label{app:vmoe}

Two public checkpoints from \citet{riquelme2021vmoe} were audited with the same probe
classes and screen in a separate implementation: V-MoE B/16 and V-MoE S/32
(exact checkpoint identifiers in App.~\ref{app:provenance}), loaded through the official
JAX/Flax implementation with no weight conversion, at the official evaluation resolution and
preprocessing (direct resize to 384, value range $[-1,1]$). Reproduced top-1 on the 50{,}000-image
validation set is $0.8554$ and $0.7578$.

$R$ is the per-MoE-layer entropy of pre-dispatch router probabilities. Capacity-constrained
dispatch can make later token representations depend on other tokens in the routing group;
pre-dispatch gates at a later layer therefore need not be independent of group composition.
The original companion check changed another routing group while preserving the anchor's
own group. Its bit-identical result establishes only that restricted invariance. No within-anchor-group intervention supplies evidence here.
Router noise is off; exact repeats check deterministic execution.
The external probe uses probability clipping $10^{-12}$ and full-array shuffles, whereas the
original local audit uses $10^{-6}$ and within-probe-fold shuffles. We retain these external
results as descriptive case studies rather than claim identical calibrated procedures.

Neither checkpoint yields a within-checkpoint three-way pass at any view under either probe
($0/8$ each). Three of
the sixteen probe--view combinations contain a standalone detection: B/16 at $O_2$ under the
linear probe (noise contrast only) and under the MLP (shuffle and noise, raw gain absent), and
S/32 at $O_3$ under the linear probe (raw gain $+0.00015$ and shuffle contrast $+0.00038$
detected, noise contrast $+0.00005$ not). None forms consistent three-way evidence. The overall
incremental-effect scale in this pretrained regime is substantially smaller than in the primary
audit, including the
positive controls, and the 10{,}000-example synthetic benchmark does not directly calibrate these
cases.

\section{Historical synthetic component-power benchmark}
\label{app:power}

\textbf{Construction.} Using real $O$ features from two representative cells, we fit a
cross-fitted $\hat\eta(O)$, draw synthetic correctness $T^\star\sim\mathrm{Bern}(\hat\eta(O))$,
and inject $R_\lambda = \lambda\,(T^\star-\hat\eta)\,e_1 + \varepsilon$ with
$\varepsilon\sim\mathcal{N}(0,I_L)$. At $\lambda=0$ the construction satisfies
$T^\star \perp R \mid O$ exactly. Each $\lambda$ is calibrated to the generator's reference information gain in
nats by a $5\times10^5$-sample Monte-Carlo evaluation of the closed-form posterior, so power
curves are indexed by a model-based reference effect size rather than by $\lambda$. Grid: true gain
$\in\{0,0.0005,0.001,0.002,0.005,0.01\}$ nats; $n\in\{500,1000,2500,5000,10000\}$; feature
dimension $L\in\{8,22\}$; 200 replicates per point; the linear probe, folds, proper scores
and component-level detection rule as the main audit.

\begin{figure}[h]
\centering
\includegraphics[width=\linewidth]{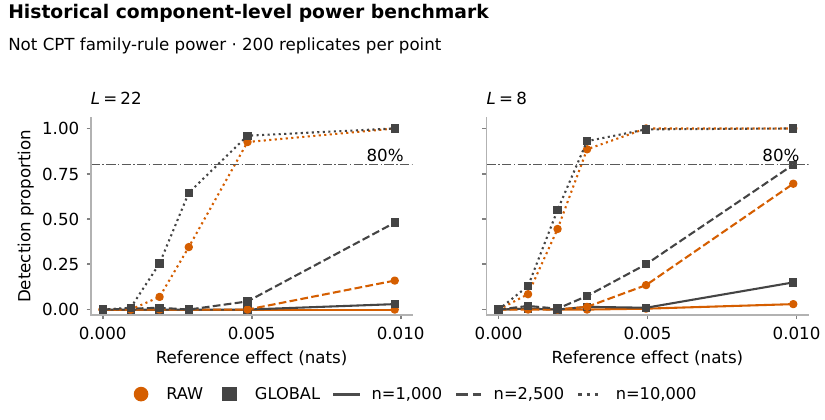}
\caption{\CapFigValidity}
\label{fig:validity}
\end{figure}

\textbf{Results} (Fig.~\ref{fig:validity}). At $n=10{,}000$ the raw-gain statistic reaches 80\%
power at $\approx$0.003--0.005 nats and the paired shuffle contrast at
$\approx$0.0027--0.0035 nats. The contrast is at least as sensitive as the raw gain in every
simulated cell at a true effect of $0.003$ nats or below, but the size of that advantage depends
on the routing dimension: at a true gain of $0.002$ nats and $n=10{,}000$ the detection rates are
$0.255$ vs $0.070$ at $L=22$ and $0.550$ vs $0.445$ at $L=8$. The two statistics converge once
both saturate, and we therefore do not claim a fixed sensitivity ratio.
Under the exact null we observed no detections in 200 replicates per setting for the
shuffle contrast and for the raw gain, and one in one setting for the noise contrast; for the
zero-event raw-gain and shuffle settings the rule-of-three 95\% upper bound is $\approx$1.5\%
per setting.

\textbf{Operating characteristics of the conjunction.} The benchmark stores
component detection rates per cell rather than the per-replicate indicators, so the joint rate
$P(D_\mathrm{raw} \wedge D_\mathrm{shuf} \wedge D_\mathrm{noise})$ is not identified by the
stored records: a joint probability is not a function of its marginals, and recovering it would
require new simulations. We instead report the assumption-free Fr\'echet bounds
$\max(0, \textstyle\sum p - 2) \le P(\text{joint}) \le \min(p_\mathrm{raw}, p_\mathrm{shuf},
p_\mathrm{noise})$ over the entire grid --- 60 cells, both feature dimensions, all six target
effects including the exact null, all five evaluation sizes, no cell excluded
(App.~\ref{app:provenance}). Two consequences follow. At zero injection the upper bound computed from the observed component proportions is
$0.000$ in all ten null cells. This is an empirical zero, not a zero upper confidence bound
on the underlying detection probability. Under the
alternative the raw-gain component is the binding minimum in 9 of the 10 cells at $n=10{,}000$,
so the \emph{upper bound} on joint component-detection power first reaches $0.8$ at $0.0049$ nats
for $L=22$ and $0.0030$ nats for $L=8$. Any actual 80\%-power threshold for the conjunction must
therefore be at least those values, which closely match the corresponding raw-gain thresholds
because raw gain is the binding marginal.
Because the benchmark applies the both-scores rule within a single replicate while the
family-level rule additionally requires agreement at $\ge$2 of 3 seeds, these are
\emph{joint component-detection} bounds, not family-level power.

\textbf{Consequence for interpretation.} The observed positive controls ($+0.0037$ to $+0.0189$
nats) lie in the powered regime. On the historical linear-probe $O_0$ slice
the largest routing contrast is $+0.0021$ nats. Larger standalone contrasts occur at other
probe--view combinations (up to $+0.0161$ nats; Table~\ref{tab:perfamily}), but none satisfies the
pre-specified three-way routing-specific criterion. \textbf{Type\,0} states that the full pre-specified
criterion was not met. Only the component non-detections are interpreted against their measured
component sensitivities; the conjunction carries Fr\'echet bounds rather than a calibrated
minimum detectable effect. No verdict asserts an effect of exactly zero.

\section{Post-hoc output-matched shuffle sensitivity analysis}
\label{app:matched}

Design fixed before implementation; this analysis is \textbf{post-hoc} and does not alter the
taxonomy of \S\ref{sec:taxonomy}. Within every training and evaluation fold we standardise $O_j$
on training-fold statistics, project to $\min(d,16)$ principal components (fit on the training
fold; needed only to make nearest-neighbour search tractable at $O_2$/$O_3$, where $d$ reaches
$200$), form blocks of $10$ by greedy nearest-neighbour grouping, and apply a random cyclic
shift inside each block. A cyclic shift is a bijection without fixed points, so the permutation
is one-to-one and self-match-free by construction. Twenty replicates per fold with fixed seeds. Each
replicate is scored on its own: out-of-fold predictions are accumulated per replicate and the
paired contrast $\Delta_b = S(\text{matched}_b) - S(\text{real})$ is formed one permutation
against one model, exactly as for the global control. We report replicate $b\!=\!1$ as the
primary comparison and use the remaining 19 only to measure permutation sensitivity. An earlier
version of this analysis averaged the 20 replicate predictions before scoring; because log loss
and Brier are convex in the prediction, that gave the control an ensemble advantage unrelated to
output matching, and it is recorded as an amendment in Table~\ref{tab:auditlog}. All other
machinery --- folds, probe classes, proper scores, paired contrasts, bootstrap,
detection rule --- is identical to the primary audit. This is an \emph{approximate output-matched
assignment shuffle}, not an exact conditional randomisation test: it substantially improves
output locality relative to the global shuffle while remaining an approximate assignment
control, particularly at the higher-dimensional views. Table~\ref{tab:matched} lists every
combination flagged by either GLOBAL draw or by MATCHED, and Table~\ref{tab:matcheddiag}
quantifies how much more local MATCHED assignments are than GLOBAL ones at each view.

\begin{table}[h]
\centering\footnotesize
\setlength{\tabcolsep}{2pt}
\caption{No matched-shuffle candidate meets the full criterion. All combinations flagged by the original GLOBAL draw, independent GLOBAL draw or MATCHED control are retained.}
\label{tab:matched}
\begin{tabular}{lllrrrcrl}
\toprule
\textbf{Family} & \textbf{Probe} & \textbf{View}
 & \multicolumn{2}{c}{\textbf{GLOBAL detected}} & \textbf{Matched} & \textbf{Matched}
 & \textbf{Reps} & \textbf{Fails} \\
 & & & audit & second & contrast & detected & & \\
\midrule
AR-block/DeiT-S/C-100 & Linear & O0 & yes & yes & $+0.0017$ & no & 11/20 & raw+noise \\
AR-block/DeiT-S/C-100 & MLP & O0 & yes & no & $+0.0015$ & no & 0/20 & raw \\
AR-block/DeiT-S/C-10 & MLP & O2 & yes & no & $+0.0036$ & no & 1/20 & noise \\
AR-full/DeiT-S/C-10 & MLP & O2 & no & no & $+0.0009$ & \textbf{yes} & 1/20 & noise \\
MoE gate/DeiT-S/C-10 & MLP & O1 & yes & no & $+0.0039$ & no & 9/20 & raw \\
AR-block/Swin-T/C-10 & MLP & O1 & yes & yes & $+0.0105$ & \textbf{yes} & 13/20 & raw+noise \\
AR-full/Swin-T/C-10 & MLP & O1 & yes & no & $-0.0093$ & no & 0/20 & raw+noise \\
AR-full/DeiT-S/T-IN & Linear & O0 & no & yes & $+0.0018$ & \textbf{yes} & 17/20 & raw \\
AR-full/DeiT-S/T-IN & MLP & O2 & no & no & $+0.0023$ & \textbf{yes} & 3/20 & raw \\
AR-full/ViT-B/C-10 & MLP & O3 & no & yes & $+0.0116$ & \textbf{yes} & 2/20 & noise \\
\bottomrule
\end{tabular}
\vspace{2pt}
\begin{minipage}{\linewidth}\footnotesize
\textit{Note.} MATCHED is replicate-1 held-out log-loss contrast (nats); Reps counts
family-rule detections across 20 permutations. Five combinations across four families are
individually detected. There are 71 family-rule flags across 3,520 replicate-level tests;
these are not independent families. The two GLOBAL draws give six and four detections,
with two in common. Fails names unmet RAW/noise components.
\end{minipage}
\end{table}

\begin{table}[h]
\centering\footnotesize
\setlength{\tabcolsep}{5pt}
\caption{MATCHED assignments are more local than GLOBAL assignments at every output view. Distances are summarised across all 22 families.}
\label{tab:matcheddiag}
\begin{tabular}{lrrrrr}
\toprule
\textbf{View} & \multicolumn{2}{c}{\textbf{MATCHED shuffle}} & \multicolumn{2}{c}{\textbf{GLOBAL shuffle}}
 & \textbf{Unmatched} \\
 & median & p95 & median & p95 & rate \\
\midrule
$O_0$ & 0.004 & 2.191 & 0.956 & 2.760 & 0.000 \\
$O_1$ & 0.744 & 4.828 & 2.695 & 5.277 & 0.000 \\
$O_2$ & 4.656 & 23.870 & 10.172 & 26.510 & 0.000 \\
$O_3$ & 8.393 & 21.561 & 12.131 & 23.150 & 0.000 \\
\bottomrule
\end{tabular}
\vspace{2pt}\begin{minipage}{\linewidth}\footnotesize
\textit{Note.} Distances use full standardised $O_j$, not the search projection, between
each example and its routing donor. Entries are medians of family summaries; p95 is the
95th percentile. No example is unmatched. Locality weakens at higher-dimensional views;
this remains an approximate assignment sensitivity check, not an exact conditional test.
\end{minipage}
\end{table}
\FloatBarrier

\section{Checkpoint-selection experiment}
\label{app:metric}

\paragraph{What was fixed in advance.} Two protocols were written before any outcome was
computed: \texttt{audit/FITTING\_METRIC\_ARM\_20260921.md} for a first pass that
re-implemented the training loop, and
\texttt{audit/FITTING\_METRIC\_INSTRUMENTED\_20260921.md} for the instrumented pass reported
here, which supersedes it. Both are post-hoc analyses relative to the original study: neither
is preregistered, and neither changes a pre-specified rule, threshold or taxonomy. The first pass is
retained in \texttt{results/fitting\_metric\_20260921}; it reproduced the direction of the
effect but not the stored count (65/120 against 57/120), which is why it was superseded.

\paragraph{The validation criterion.} The nonlinear probe is
\begin{flushleft}\small\quad\texttt{MLPClassifier(hidden\_layer\_sizes=(64,), alpha=1e-3, max\_iter=300,}\\
\quad\texttt{early\_stopping=True, n\_iter\_no\_change=15)}.\end{flushleft}
Early stopping is off by default in scikit-learn;
when it is enabled, as here, 10\% of the training fold is held out (the default
\texttt{validation\_fraction}), \texttt{\_update\_no\_improvement\_count} scores that split with
\texttt{\_score}, which for a classifier is \texttt{accuracy\_score} on hard predictions, and the
weights restored at the end of \texttt{\_fit\_stochastic} are those with the best validation
accuracy. The training loss is the $L_2$-penalised log loss throughout. Every comparison in this
paper is scored by log loss and Brier score.

\paragraph{Instrumentation.} \texttt{InstrumentedMLP} subclasses \texttt{MLPClassifier} and
overrides \texttt{\_score} only. The accuracy value is taken from \texttt{super().\_score},
so it is the stock quantity; the log loss is computed separately from probabilities via
\texttt{\_forward\_pass\_fast}, never by substituting \texttt{log\_loss} into the hard-label
scoring helper. \texttt{\_score} consumes no random state, and arm~A is verified against the
stock estimator on all six cells: maximum absolute difference in out-of-fold probabilities
$0.0$, with identical \texttt{n\_iter\_}.

\paragraph{Arms.} All share hidden width 64, $\alpha=10^{-3}$, \texttt{random\_state=0}, Adam,
the stored folds, standardisation, labels and bootstrap draws.
\emph{A} returns the stock accuracy score for both stopping and selection and equals the original
procedure. \emph{B} returns the negative validation log loss (scikit-learn maximises the validation
score), which changes the whole stopping-and-selection policy, so its effect may act through
training length as well as selection. \emph{C} disables early stopping, trains every fit for exactly
100 epochs and reads the one recorded trajectory twice, selecting by validation accuracy
(\emph{C\textsubscript{acc}}) or by validation log loss (\emph{C\textsubscript{nll}}) on the inner
split only. The outer test fold never chooses an epoch. Comparing C\textsubscript{acc} with
C\textsubscript{nll} isolates the selection criterion at a fixed trajectory; comparing A with B
does not. The selected epochs in Tables~\ref{tab:metric} and~\ref{tab:torchsel} are means over
fits, first within each record and then over records; they are not the peak of any single
trajectory, and in the C arms a checkpoint selected early does not mean that training stopped
early.

\paragraph{Results.} Table~\ref{tab:t2summary} records which check licenses which claim;
Table~\ref{tab:metric} pools the null panel, Table~\ref{tab:metriccells} gives each cell, and
Table~\ref{tab:poscontrol} the implanted-signal control. The protocol fixed three readings in
advance: selection criterion alone (C\textsubscript{nll} far below C\textsubscript{acc} at a fixed
trajectory), training length (C\textsubscript{acc} close to C\textsubscript{nll} while B differs from
A), or no reproduction under instrumentation. The result is the first: the selection criterion
accounts for the effect, since C\textsubscript{acc} and C\textsubscript{nll} share a trajectory. The
detections are concentrated in some models (Table~\ref{tab:metriccells}), so the pooled count
describes this six-model panel rather than a population of models.

\paragraph{Independent implementation.} A separate post-hoc protocol
(\path{audit/TORCH_SELECTION_PROTOCOL_20260925.md}), written before any of its outcomes were
computed, repeats the C design with a PyTorch~1.13.1 MLP: one hidden layer of 64 ReLU units,
binary cross-entropy plus the penalty $\tfrac12\alpha\lVert W\rVert^2/\text{batch}$ on weights only
(scikit-learn's definition, $\alpha=10^{-3}$), Adam with learning rate $10^{-3}$, batch size 200,
a stratified 10\% validation split of each training fold, float32 arithmetic and PyTorch's default
initialisation. Each fit runs exactly 100 epochs and is read at the first epoch with maximal
validation accuracy, at the first epoch with minimal validation log loss and at the final epoch.
Cells, label draws, implanted signals, folds, standardisation, clipping, bootstrap draws and the
flag rule are those of the scikit-learn panel, so records pair one to one. A further null mode
replaces routing by independent standard-normal features of the same width with the same labels.
Table~\ref{tab:torchsel} reports all six cells. The pre-specified reading ``reproduced'' (at least
10/120 accuracy-selected flags and at most 6/120 log-loss-selected flags) holds for both null modes.
The cells that fail differ from the scikit-learn panel, and the overall rate is higher, which we
attribute to the remaining implementation differences without having isolated them.

\begin{table}[htbp]\centering\small\setlength{\tabcolsep}{4pt}
\caption{Instrumented scikit-learn MLP under the label null: six cells $\times$ 20 label draws at
$O_0$; only the validation score changes. Arm A reproduces the stored estimator bit for bit; the
C arms share one 100-epoch trajectory per fit and differ only in the checkpoint selected on the
inner validation split. Gain is the mean fitted log-loss contrast; ``vs $g$'' is the mean held-out
loss minus the frozen generator's loss on the same label draw (a diagnostic, not a population
excess risk); selected and run epochs are means over fits. Flags count the 120 draws of this fixed
panel, which are not independent draws of models (Table~\ref{tab:metriccells}).}
\label{tab:metric}
\begin{tabular}{lrrrrrr}\toprule
Arm & Flags & Gain & \multicolumn{2}{c}{vs $g$ (nats)} & Selected ep. & Run \\
\cmidrule(lr){4-5}
 & & (nats) & output & $+$R & out / $+$R & ep. \\\midrule
A: accuracy stopping and selection (stock) & 57/120 & +0.0169 & +0.0365 & +0.0196 & 11 / 19 & 27 \\
B: log-loss stopping and selection & 0/120 & -0.0071 & +0.0008 & +0.0079 & 35 / 40 & 45 \\
C\textsubscript{acc}: 100 epochs, accuracy checkpoint & 50/120 & +0.0119 & +0.0264 & +0.0144 & 18 / 37 & 100 \\
C\textsubscript{nll}: 100 epochs, log-loss checkpoint & 0/120 & -0.0073 & +0.0007 & +0.0080 & 56 / 42 & 100\\
\bottomrule\end{tabular}\end{table}

\begin{table}[htbp]\centering\footnotesize\setlength{\tabcolsep}{3.5pt}
\caption{Checkpoint-selection arms, per cell. Twenty label draws per cell at $O_0$ with the
MLP probe. Flagged counts are raw detections under arm A (original accuracy policy), B (log-loss
policy) and the two fixed-trajectory arms C\textsubscript{acc} and C\textsubscript{nll}; the
excess columns are the fixed-trajectory arms' mean held-out log loss of the output-only probe
minus the frozen generator's loss on the same labels.}\label{tab:metriccells}
\begin{tabular}{lrrrrrr}\toprule
& \multicolumn{4}{c}{Flagged (of 20)} & \multicolumn{2}{c}{Output-only excess vs $g$ (nats)} \\
\cmidrule(lr){2-5}\cmidrule(lr){6-7}
Cell & A & B & C\textsubscript{acc} & C\textsubscript{nll} & C\textsubscript{acc} & C\textsubscript{nll} \\\midrule
MoE/DeiT/C-10 & 7/20 & 0/20 & 4/20 & 0/20 & +0.0122 & +0.0005 \\
AR-block/Swin/C-10 & 1/20 & 0/20 & 2/20 & 0/20 & +0.0141 & +0.0005 \\
AR-block/DeiT/C-100 & 6/20 & 0/20 & 4/20 & 0/20 & +0.0144 & +0.0009 \\
Halting/DeiT/C-100 & 20/20 & 0/20 & 20/20 & 0/20 & +0.0630 & +0.0008 \\
AR-block/ViT/C-100 & 3/20 & 0/20 & 1/20 & 0/20 & +0.0160 & +0.0006 \\
AR-full/DeiT/TIN & 20/20 & 0/20 & 19/20 & 0/20 & +0.0384 & +0.0010\\
\bottomrule\end{tabular}\end{table}

\begin{table}[htbp]\centering\footnotesize\setlength{\tabcolsep}{4pt}
\caption{Checkpoint selection on fixed 100-epoch trajectories in two implementations, view $O_0$,
MLP probe. Entries are raw-comparison flags (both bootstrap intervals above zero). scikit-learn:
the instrumented stock estimator, arms C\textsubscript{acc} and C\textsubscript{nll}
(Table~\ref{tab:metric}). PyTorch: an independently implemented MLP with the same width,
penalty definition, optimiser settings, folds, labels and scoring, read at the checkpoint with
the best validation accuracy (acc), the lowest validation log loss (nll) or the final epoch
(last). Null: labels from the frozen output-only generator with real routing (20 draws per
cell) or with routing replaced by independent Gaussian features of the same width (Gaussian
routing, same labels). Implanted signal: 10 draws per cell and strength.}
\label{tab:torchsel}
\begin{tabular}{lrrrrrrrr}\toprule
 & \multicolumn{2}{c}{\shortstack{scikit-learn\\null}} & \multicolumn{3}{c}{\shortstack{PyTorch\\null}} & \multicolumn{3}{c}{\shortstack{PyTorch, null with\\Gaussian routing}} \\
\cmidrule(lr){2-3}\cmidrule(lr){4-6}\cmidrule(lr){7-9}
Cell & acc & nll & acc & nll & last & acc & nll & last \\\midrule
AR-block/DeiT/C-100 & 4 & 0 & 20 & 0 & 0 & 19 & 0 & 0 \\
AR-block/Swin/C-10 & 2 & 0 & 0 & 0 & 0 & 4 & 0 & 0 \\
AR-full/DeiT/TIN & 19 & 0 & 18 & 0 & 0 & 14 & 0 & 0 \\
AR-block/ViT/C-100 & 1 & 0 & 20 & 0 & 0 & 19 & 0 & 0 \\
MoE/DeiT/C-10 & 4 & 0 & 5 & 0 & 0 & 4 & 0 & 0 \\
Halting/DeiT/C-100 & 20 & 0 & 20 & 0 & 0 & 20 & 0 & 0 \\
\midrule All six (of 120) & 50 & 0 & 83 & 0 & 0 & 80 & 0 & 0 \\\midrule
 & \multicolumn{2}{c}{scikit-learn} & \multicolumn{3}{c}{PyTorch} & & & \\
Implanted signal $\approx$0.005 nats (of 60) & 20 & 0 & 47 & 1 & 0 & & &  \\
Implanted signal $\approx$0.01 nats (of 60) & 27 & 2 & 46 & 5 & 1 & & & \\
\bottomrule\end{tabular}\end{table}

\begin{table}[htbp]
\centering\footnotesize
\setlength{\tabcolsep}{3pt}
\caption{\textbf{Validation checks and observed-label evidence establish different claims.}
The common-generator benchmark assesses null detections, under the original accuracy-based
checkpoint selection and, for the MLP half, under log-loss selection; the independent-split
benchmark assesses a linear power endpoint. R1 and R2 report distinct evidential
criteria.}\label{tab:t2summary}
\begin{tabular}{>{\raggedright\arraybackslash}p{.23\linewidth}>{\raggedright\arraybackslash}p{.22\linewidth}>{\raggedright\arraybackslash}p{.23\linewidth}>{\raggedright\arraybackslash}p{.24\linewidth}}
\toprule
Evidence / check & Setting & Result & Interpretation \\
\midrule
\multicolumn{4}{l}{\textit{Validation}} \\
Exact-label-null CPT & Common-generator benchmark & 59/3,840 (1.54\%) & All pooled gates pass; max cell 4\% \\
Null comparator detections & RAW / GLOBAL / MATCHED & 528 / 325 / 305 of 3,840 & Contrast concentrated in MLP \\
Same, log-loss selection & MLP half re-run, paired & 0 / 27 / 26 of 1,920; CPT 22 & Selection rule, not function class \\
Power adequacy & Independent 5k/5k split & 200/200 + 200/200 & Linear endpoint only \\
\midrule
\multicolumn{4}{l}{\textit{Observed-label evidence}} \\
R1: conditional assignment & Original held-out labels & 12/176 cells; 5 families & Model-relative assignment evidence \\
R2: same-width robustness & Same test rows + noise & 0/176 cells; 0 families & Additional robustness unmet \\
\bottomrule
\end{tabular}
\vspace{2pt}
\begin{minipage}{\linewidth}\footnotesize
\textit{Note.} Null counts are replicate--probe evaluations on 5k test examples sharing label
vectors; R1/R2 count family--probe--view cells with at least two positive seeds. Rows without
a stated policy are accuracy-policy results, the behaviour of the originally defined
procedure. Exactness concerns labels, not estimated-$q$ CPT validity.
\end{minipage}
\end{table}

\begin{table}[htbp]\centering\small\setlength{\tabcolsep}{4pt}
\caption{Table~\ref{tab:nllcommon} resolved by output view. Views share label draws, so rows
within a comparator are not independent.}
\label{tab:nllcommonfull}
\begin{tabular}{llrrrrrrr}\toprule
Comparator & View & $n$ & $n_{00}$ & $n_{01}$ & $n_{10}$ & $n_{11}$ & acc (\%) & nll (\%) \\\midrule
RAW & $O_0$ & 600 & 292 & 0 & 308 & 0 & 51.33 & 0.00 \\
 & $O_1$ & 600 & 508 & 0 & 92 & 0 & 15.33 & 0.00 \\
 & $O_2$ & 360 & 303 & 0 & 57 & 0 & 15.83 & 0.00 \\
 & $O_3$ & 360 & 289 & 0 & 71 & 0 & 19.72 & 0.00 \\
\cmidrule(lr){2-9} & all & 1920 & 1392 & 0 & 528 & 0 & 27.50 & 0.00 \\
GLOBAL & $O_0$ & 600 & 479 & 2 & 112 & 7 & 19.83 & 1.50 \\
 & $O_1$ & 600 & 513 & 2 & 84 & 1 & 14.17 & 0.50 \\
 & $O_2$ & 360 & 303 & 5 & 52 & 0 & 14.44 & 1.39 \\
 & $O_3$ & 360 & 303 & 5 & 47 & 5 & 14.44 & 2.78 \\
\cmidrule(lr){2-9} & all & 1920 & 1598 & 14 & 295 & 13 & 16.04 & 1.41 \\
MATCHED & $O_0$ & 600 & 507 & 2 & 88 & 3 & 15.17 & 0.83 \\
 & $O_1$ & 600 & 503 & 3 & 93 & 1 & 15.67 & 0.67 \\
 & $O_2$ & 360 & 305 & 2 & 50 & 3 & 14.72 & 1.39 \\
 & $O_3$ & 360 & 302 & 3 & 46 & 9 & 15.28 & 3.33 \\
\cmidrule(lr){2-9} & all & 1920 & 1617 & 10 & 277 & 16 & 15.26 & 1.35 \\
CPT & $O_0$ & 600 & 585 & 5 & 10 & 0 & 1.67 & 0.83 \\
 & $O_1$ & 600 & 586 & 4 & 9 & 1 & 1.67 & 0.83 \\
 & $O_2$ & 360 & 349 & 3 & 6 & 2 & 2.22 & 1.39 \\
 & $O_3$ & 360 & 352 & 6 & 1 & 1 & 0.56 & 1.94 \\
\cmidrule(lr){2-9} & all & 1920 & 1872 & 18 & 26 & 4 & 1.56 & 1.15\\
\bottomrule\end{tabular}\end{table}

\begin{table}[htbp]\centering\small
\caption{Observed-label MLP tests recomputed under log-loss checkpoint selection with the same conditional model, split and reference assignments, for all 68 runs (272 run--view tests). Seed-level counts are single tests. R1 is positive in 2 and 2 of 88 family--view cells under the two policies, but not in the same cells: AR-block/DeiT/C-100 at $O_0$ becomes R1-positive; AR-full/DeiT/C-100 at $O_2$ loses R1. Post-hoc re-analysis; the accuracy-policy results are those of the pre-specified procedure.}
\label{tab:nllobserved}
\begin{tabular}{lrr}\toprule
 & Accuracy policy & Log-loss policy \\\midrule
Seed-level CPT flags (recomputed runs) & 19/272 & 21/272 \\
\quad both / accuracy only / log-loss only / neither & \multicolumn{2}{c}{9 / 10 / 12 / 241} \\
R1 cells (family--view) & 2/88 & 2/88 \\
Mean conditional contrast, recomputed runs (nats) & +0.000379 & +0.001328 \\
\bottomrule\end{tabular}\end{table}

\begin{table}[htbp]\centering\footnotesize\setlength{\tabcolsep}{3pt}
\caption{Matched synthetic panel at view $O_1$, MLP probe: the first 20 replicates of each setting
of the independent-split benchmark (Table~\ref{tab:power}) analysed under the original
accuracy-based stopping-and-selection policy and under the log-loss policy, sharing labels,
routing, folds, controls, conditional model and reference assignments. Counts are per-replicate
flags out of 20; the mean raw gain is the average fitted log-loss contrast over the 20
replicates. The reference gain is a property of the implanted synthetic signal, not an observed
conditional-reference contrast. CPT $\wedge$ noise is a per-replicate conjunction, not the
cross-seed R2 rule. Twenty paired replicates give coarse resolution.}
\label{tab:nllpanel}
\begin{tabular}{lllrrrrrr}\toprule
Cell & Ref.\ gain & Selection & RAW & Mean RAW & GLOBAL & MATCHED & CPT & CPT $\wedge$ \\
 & (nats) & & flags & gain (nats) & flags & flags & flags & noise \\\midrule
Swin/C-10 & 0 & accuracy & 11/20 & $+0.0267$ & 7/20 & 7/20 & 0/20 & 0/20 \\
 &  & log loss & 0/20 & $-0.0090$ & 0/20 & 0/20 & 0/20 & 0/20 \\
Swin/C-10 & 0.0049 & accuracy & 16/20 & $+0.0501$ & 13/20 & 11/20 & 3/20 & 3/20 \\
 &  & log loss & 0/20 & $-0.0040$ & 4/20 & 3/20 & 11/20 & 6/20 \\
Swin/C-10 & 0.0098 & accuracy & 16/20 & $+0.0654$ & 18/20 & 16/20 & 10/20 & 9/20 \\
 &  & log loss & 1/20 & $+0.0009$ & 19/20 & 15/20 & 20/20 & 15/20 \\
DeiT/C-100 & 0 & accuracy & 0/20 & $-0.0104$ & 0/20 & 0/20 & 0/20 & 0/20 \\
 &  & log loss & 0/20 & $-0.0126$ & 0/20 & 0/20 & 1/20 & 0/20 \\
DeiT/C-100 & 0.0051 & accuracy & 0/20 & $-0.0058$ & 3/20 & 5/20 & 7/20 & 3/20 \\
 &  & log loss & 0/20 & $-0.0078$ & 4/20 & 5/20 & 10/20 & 3/20 \\
DeiT/C-100 & 0.0102 & accuracy & 2/20 & $-0.0005$ & 14/20 & 15/20 & 18/20 & 16/20 \\
 &  & log loss & 0/20 & $-0.0027$ & 12/20 & 16/20 & 19/20 & 14/20\\
\bottomrule\end{tabular}\end{table}

\begin{table}[htbp]\centering\small
\caption{Positive control with the same pipeline: labels and a synthetic routing matrix
$R_\lambda=\lambda(T-\eta)e_1+\varepsilon$ carrying a known conditional signal, six cells
$\times$ 10 replicates per strength. Log-loss selection removes the null detections of
Table~\ref{tab:metric} and also detects almost nothing here, while accuracy selection flags at
close to its null rate either way. The last column shows that the augmented probe remains
worse than the generator even where real signal exists.}
\label{tab:poscontrol}
\begin{tabular}{llrr}\toprule
Reference gain (nats) & Arm & Flagged & $+$R vs $g$ (nats) \\\midrule
0.0050 & A: accuracy stopping and selection (stock) & 29/60 & +0.0163 \\
0.0050 & B: log-loss stopping and selection & 0/60 & +0.0069 \\
0.0050 & C\textsubscript{acc}: 100 epochs, accuracy checkpoint & 20/60 & +0.0141 \\
0.0050 & C\textsubscript{nll}: 100 epochs, log-loss checkpoint & 0/60 & +0.0069 \\
0.0101 & A: accuracy stopping and selection (stock) & 37/60 & +0.0107 \\
0.0101 & B: log-loss stopping and selection & 2/60 & +0.0020 \\
0.0101 & C\textsubscript{acc}: 100 epochs, accuracy checkpoint & 27/60 & +0.0084 \\
0.0101 & C\textsubscript{nll}: 100 epochs, log-loss checkpoint & 2/60 & +0.0020\\
\bottomrule\end{tabular}\end{table}

\FloatBarrier

\paragraph{Reading the generator comparison.} The ``vs $g$'' columns are empirical held-out
losses minus the empirical loss of the frozen generator on the same realised labels, a diagnostic
on one label draw rather than a population excess risk; on a finite sample the generator's
empirical loss need not be the minimum attainable. The generator never selects a checkpoint,
fold or replicate. A second, fixed-design diagnostic holds covariates and fitted predictions fixed
and takes expectations over a fresh label draw from the generator. For the C arms the output-only
probe then carries 0.0264 nats of excess log loss under accuracy selection and 0.0009 under
log-loss selection, and the routing-augmented probe 0.0146 and 0.0081. Because the incremental
information is zero by construction, the identity fitted gain $=$ baseline excess $-$ augmented
excess gives $+0.0118$ and $-0.0073$ nats (from unrounded values), within $2\times10^{-4}$ of the
measured mean gains of $+0.0119$ and $-0.0073$. The identity is an accounting check, not an
independent proof of the mechanism.

\paragraph{What this does not show.} The instrumented experiments cover $O_0$ and the MLP probe on
six cells in two implementations. The benchmark-wide re-analysis
(Tables~\ref{tab:nllcommonfull}--\ref{tab:nllpanel}) uses the B-type log-loss policy for all 1,920
common-generator MLP evaluations, the matched panel and all 68 observed-label runs, while the
pre-specified screen and all linear evaluations keep the original rule. The fixed-trajectory
intervention isolates the padded-baseline mechanism for the raw comparison only; it does not
establish that every positive elsewhere in the benchmark has this cause. Changing the probe's
selection rule inside a CPT requires recomputing the observed statistic and all 79 reference
statistics of every test, which the re-analysis does.

\section{Common-generator exact-label-null validation}
\label{app:t4}

\begin{figure}[htbp]
\centering\includegraphics[width=0.86\linewidth]{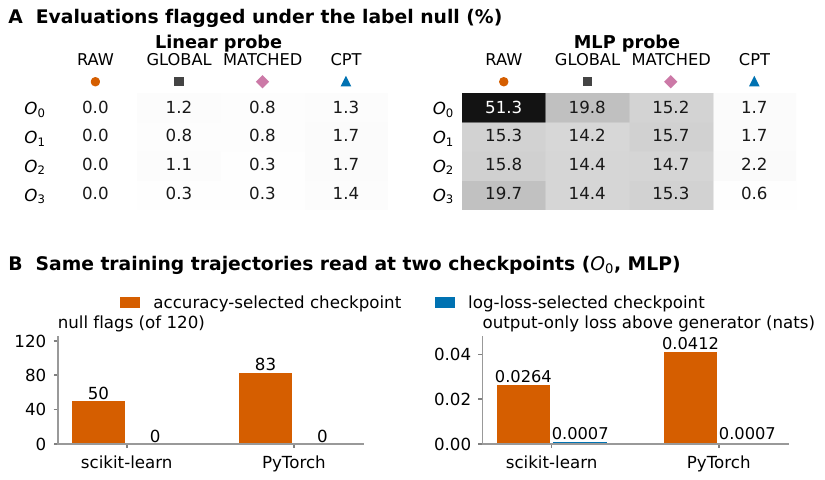}
\caption{\CapFigMech}\label{fig:mech}
\end{figure}

The common-generator exact-label-null benchmark supplies the primary null-validity evidence.
Its design and interpretation gates were pre-specified. It keeps
the six historical model seeds, representations, probes, conditional-model fitting, scores,
and CPT settings fixed while replacing the fold-specific label recipe with one common
generator per model. The historical benchmark remains in App.~\ref{app:exactnull} and
Table~\ref{tab:t2nullcontrols}; neither its counts nor the original taxonomy is changed.
Figure~\ref{fig:mech} summarises the detections of \S\ref{sec:exactnull} by view and the
fixed-trajectory checkpoint experiment of \S\ref{sec:mechanism}.

\subsection{Construction, exactness and implementation checks}

Use the sorted seed-204 disjoint 5k/5k split. On A alone fit
$g_m(o)=\sigma(\beta_{0,m}+\beta_{1,m}o)$ to scalar $O_0$ and observed correctness.
The fixed logistic constructor uses L2 penalty, C=1, lbfgs, intercept, tolerance $10^{-4}$,
a maximum of 3000 iterations and random seed 0; no scaling, class weights or tuning is added.
Freeze the model and its ordered $B_{\rm test}$ probability vector before generating labels.
For each model, save 100 vectors using independent-uniform label draws in the ideal model.
Replicates 0--59 are reused at all four views; 60--99 at O0/O1 only. Both probes read the same
stored vector. Thus 600 unique vectors produce 1,920 model--view--replicate records and
3,840 replicate--probe evaluations per comparison, not 3,840 independent label draws.

Conditional on A, $g_A$ is fixed and
\[
P(T^*=1\mid O_0,R,A)=g_A(O_0),\qquad
T^*\perp R\mid O_j,A\quad\text{if }O_0=h_j(O_j).
\]
O1 contains O0; O2 recovers its clipped logit from the largest probability; O3 recovers it
from the softmax of sorted logits. In stored arithmetic O0/O1/O2 recover exactly. O3's
maximum discrepancies across models are $5.8\times10^{-15}$ to $2.1\times10^{-13}$;
no identical stored $O_j$ rows have different O0 values, so the factorisation holds on the
evaluated support. Identical O0 rows also have identical generator probabilities.
The broader unconditional statement needs the i.i.d.-split assumptions, not just disjoint
indices. Independent uniforms are the mathematical assumption; distinct deterministic PRNG
seeds prevent identical-stream reuse but do not prove independence of pseudorandom streams.
Shared labels and the common fitted A induce dependence across reported analyses.

Label-generation streams use a dedicated, non-overlapping deterministic namespace
(App.~\ref{app:randomstreams}). All 600 seeds are distinct and disjoint from the
control/bootstrap/CPT namespaces. Conditional-model fitting uses A only; every probe
and test statistic uses B only. All records use 79 spokes and S=50, with identical
labels reused across eligible output views and probes.

\begin{table}[htbp]
\centering\footnotesize
\setlength{\tabcolsep}{3pt}
\caption{A single fitted generator per model defines the label benchmark. Coefficients use A; descriptive probability and score diagnostics use B.}\label{tab:t4generators}
\begin{tabular}{lrrrrrrr}
\toprule
Model & $\beta_0$ & $\beta_1$ & Mean p & SD p & Observed $T$ & Log loss & Brier \\
\midrule
MoE/DeiT/C10 & 0.807199 & 1.972213 & 0.9228 & 0.1354 & 0.9224 & 0.1840 & 0.0534 \\
ARb/Swin/C10 & 0.989538 & 1.882043 & 0.9035 & 0.1605 & 0.9042 & 0.2156 & 0.0665 \\
ARb/DeiT/C100 & 0.850323 & 1.240079 & 0.6767 & 0.2773 & 0.6922 & 0.4312 & 0.1416 \\
ACT/DeiT/C100 & 0.909690 & 1.316111 & 0.7346 & 0.2551 & 0.7356 & 0.4072 & 0.1322 \\
ARb/ViT/C100 & 0.752829 & 1.321237 & 0.7106 & 0.2729 & 0.7234 & 0.4102 & 0.1336 \\
ARf/DeiT/TIN & 0.568209 & 1.046911 & 0.5468 & 0.2906 & 0.5390 & 0.5082 & 0.1689 \\
\bottomrule
\end{tabular}
\vspace{2pt}
\begin{minipage}{\linewidth}\footnotesize
\textit{Note.} All six fits converged without warnings; no diagnostic selected a generator or changed the gate. Mean/SD $p$ describe generated probabilities; observed $T$ is correctness prevalence.
\end{minipage}
\end{table}

Generator probabilities are not clipped; only proper-score evaluation uses the frozen
$[10^{-6},1-10^{-6}]$ clip. Table~\ref{tab:t4generators} separates A-fit coefficients from B
diagnostics. Probability extrema, all five frozen quantiles, constructors, fitted states,
warnings are retained in the numerical supplement (App.~\ref{app:provenance}).
No diagnostic served as a selection or stopping threshold.

\subsection{Complete calibration and historical comparison}

The operational gate requires 6--24 positives out of 600 at O0/O1 and 2--16 out of 360 at
O2/O3, separately for each probe, plus no individual model--view--probe cell above 10\%.
All eight pooled bands and all 48 cell ceilings pass. The total is 59/3,840=1.536458\%;
the maximum cell is 4/100=4\%. O3/MLP is at the pooled lower boundary (2/360), with both
positives in ViT/C100 (2/60). These are empirical operational gates, not a joint confidence
guarantee or proof that estimated-$q$ CPT has exact nominal size. Table~\ref{tab:t4comparison}
compares historical fold-specific and common-generator detections by view and probe, and
Table~\ref{tab:t4cells} lists all 48 model--view--probe CPT cells against the 10\% ceiling.

\begin{table}[htbp]
\centering\footnotesize
\setlength{\tabcolsep}{3pt}
\caption{The nonlinear contrast persists across label generators, while local comparator ordering changes. Entries compare historical fold-specific $\to$ common-generator detections.}\label{tab:t4comparison}
\begin{tabular}{llrrrrr}
\toprule
View & Probe & Denominator & RAW & GLOBAL & MATCHED & CPT \\
\midrule
O0 & Linear & 600 & 0$\to$0 & 2$\to$7 & 5$\to$5 & 12$\to$8 \\
O0 & MLP & 600 & 286$\to$308 & 109$\to$119 & 92$\to$91 & 13$\to$10 \\
O1 & Linear & 600 & 0$\to$0 & 3$\to$5 & 3$\to$5 & 9$\to$10 \\
O1 & MLP & 600 & 102$\to$92 & 93$\to$85 & 85$\to$94 & 14$\to$10 \\
O2 & Linear & 360 & 0$\to$0 & 3$\to$4 & 2$\to$1 & 7$\to$6 \\
O2 & MLP & 360 & 59$\to$57 & 63$\to$52 & 67$\to$53 & 6$\to$8 \\
O3 & Linear & 360 & 0$\to$0 & 4$\to$1 & 1$\to$1 & 9$\to$5 \\
O3 & MLP & 360 & 63$\to$71 & 53$\to$52 & 37$\to$55 & 4$\to$2 \\
\bottomrule
\end{tabular}
\vspace{2pt}
\begin{minipage}{\linewidth}\footnotesize
\textit{Note.} Each entry uses the row denominator in replicate--probe evaluations and requires both proper scores. Shared label vectors induce dependence. These are descriptive counts, not new significance tests.
\end{minipage}
\end{table}

\begin{table}[htbp]
\centering\footnotesize
\setlength{\tabcolsep}{3pt}
\caption{Every common-generator CPT cell meets the pre-specified ceiling. All 48 model--view--probe cells are shown.}\label{tab:t4cells}
\begin{tabular}{llrrrr}
\toprule
Model & Probe & O0 & O1 & O2 & O3 \\
\midrule
MoE/DeiT/C10 & Linear & 2/100 (2.00\%) & 3/100 (3.00\%) & 1/60 (1.67\%) & 1/60 (1.67\%) \\
MoE/DeiT/C10 & MLP & 1/100 (1.00\%) & 0/100 (0.00\%) & 2/60 (3.33\%) & 0/60 (0.00\%) \\
ARb/Swin/C10 & Linear & 1/100 (1.00\%) & 1/100 (1.00\%) & 1/60 (1.67\%) & 0/60 (0.00\%) \\
ARb/Swin/C10 & MLP & 1/100 (1.00\%) & 2/100 (2.00\%) & 1/60 (1.67\%) & 0/60 (0.00\%) \\
ARb/DeiT/C100 & Linear & 1/100 (1.00\%) & 1/100 (1.00\%) & 1/60 (1.67\%) & 2/60 (3.33\%) \\
ARb/DeiT/C100 & MLP & 2/100 (2.00\%) & 4/100 (4.00\%) & 1/60 (1.67\%) & 0/60 (0.00\%) \\
ACT/DeiT/C100 & Linear & 3/100 (3.00\%) & 2/100 (2.00\%) & 1/60 (1.67\%) & 2/60 (3.33\%) \\
ACT/DeiT/C100 & MLP & 2/100 (2.00\%) & 2/100 (2.00\%) & 2/60 (3.33\%) & 0/60 (0.00\%) \\
ARb/ViT/C100 & Linear & 0/100 (0.00\%) & 1/100 (1.00\%) & 0/60 (0.00\%) & 0/60 (0.00\%) \\
ARb/ViT/C100 & MLP & 3/100 (3.00\%) & 1/100 (1.00\%) & 1/60 (1.67\%) & 2/60 (3.33\%) \\
ARf/DeiT/TIN & Linear & 1/100 (1.00\%) & 2/100 (2.00\%) & 2/60 (3.33\%) & 0/60 (0.00\%) \\
ARf/DeiT/TIN & MLP & 1/100 (1.00\%) & 1/100 (1.00\%) & 1/60 (1.67\%) & 0/60 (0.00\%) \\
\bottomrule
\end{tabular}
\vspace{2pt}
\begin{minipage}{\linewidth}\footnotesize
\textit{Note.} Entries are detections/evaluations (percent); all six models use training seed 0. The maximum is ARb/DeiT/C100, O1/MLP, 4/100=4\%; no cell exceeds 10\%. Exactness concerns the label null, not estimated-$q$ CPT.
\end{minipage}
\end{table}

The historical totals RAW/GLOBAL/MATCHED/CPT are 510/330/292/74, each out of 3,840;
the common-generator benchmark gives 528/325/305/59, each out of 3,840 replicate--probe evaluations. The broad nonlinear CPT contrast persists, but comparator ordering
is not invariant: at O1/MLP, historical RAW$>$GLOBAL$>$MATCHED becomes
MATCHED$>$RAW$>$GLOBAL; at O2/MLP, MATCHED$>$GLOBAL$>$RAW becomes
RAW$>$MATCHED$>$GLOBAL. These are descriptive comparisons with no post-hoc significance test.
Linear comparators are also low, including zero RAW detections. The common-generator benchmark resolves the label-generator objection but
does not retroactively make the historical construction exact. Its q estimate may still be
misspecified, and empirical seed-level calibration does not calibrate a replicated R1/R2
family rule or establish corrected MLP/full-curve/R2 power.

\Needspace{8\baselineskip}
\section{Conditional-permutation analysis: evidence and validation}
\label{app:cpt}

The conditional extension is a secondary analysis of the Routing Information Audit (RIA). Its
primary arm uses independent halves for fitting $\hat q$ and for the CPT statistic; it does not
replace the pre-specified 10k global-shuffle screen. The common-generator benchmark is reported in
App.~\ref{app:t4}; methodological corrections and excluded analyses are documented
in App.~\ref{app:provenance}.

\subsection{Working conditional model and assignment weights}

Use a fixed seed-204 random permutation of the 10,000 examples for each run and view.
Sort the first 5,000 indices for A and sort the
remaining indices for $B_{\rm test}$. q selection, scaling, empirical routing ranks and
covariance fitting use A only, without correctness labels. Only routing rows from
$B_{\rm test}$ are permuted; all outer probe training and evaluation folds stay within that
half. Probes use five correctness-stratified folds, two repeats at seeds 100/101, with each
row's two held-out predictions averaged before proper scoring.

The working model transforms each routing coordinate by its A-sample empirical rank to a
normal quantile, with linear interpolation and tail clamping. O is standardised on A; no
PCA is applied to O. Q1 uses a ridge mean, Q2 a gradient-boosted mean, with shared covariance;
Q3 permits output-dependent diagonal scale with shared correlation. A Ledoit--Wolf estimate
regularises the residual correlation. Unstratified three-fold inner selection (seed 203)
maximises held-out conditional log likelihood entirely inside A. Ridge mean penalties are
$\{10^{-3},10^{-2},10^{-1},1,10,100\}$; scale penalties are $\{10^{-1},1,10,100\}$.
The boosting grid fixes 200 iterations, learning rate 0.05, minimum leaf size 50 and
15 or 31 leaves. Q3 must improve on the best Q1/Q2 held-out likelihood by more than 1.0 nat
per example. All 272 primary run--view fits select Q1. This describes an estimated working
assignment model, not a verified Gaussian law for bounded routing entropies.

For $q(r\mid o)=h(r)\exp\{\eta(o)^\top T(r)-A(o)\}$, permutation weights satisfy
\[
\prod_i q(R_{\pi(i)}\mid O_i)\ \propto\
\exp\!\left\{\sum_i\eta(O_i)^\top T(R_{\pi(i)})\right\}.
\]
The base measure and normaliser cancel because the routing multiset and outputs are fixed.
For a shared-covariance Gaussian model on transformed $z(R)$, the surviving term is
$\hat m(O_i)^\top\hat\Sigma^{-1}z(R_{\pi(i)})$. Output-dependent diagonal scale introduces
additional quadratic terms. This cancellation does not remove misspecification of the
output-dependent routing structure, including correlations and higher-order shape.

\subsection{Conditional sampler and probe settings}

Each transition randomly pairs positions and performs reversible heat-bath swaps using the
conditional log-weight difference. The observed assignment runs 50 rounds to a hub; each
of 79 independent spokes then runs 50 rounds from that hub. With the true conditional law,
this construction gives exchangeability with the observed assignment for fixed S
\citep{berrett2020conditional}. An estimated law can be misspecified even when fitted on
independent examples. Movement diagnostics do not certify conditional fidelity.

Independent deterministic random streams are assigned to model, output view, transition
leg and replicate (App.~\ref{app:randomstreams}). Conditional-model dimensionality is
family-specific (Table~\ref{tab:t2dimensions}).

Linear probes use logistic regression with C=1 and a maximum of 3000 iterations. MLPs use one 64-unit
hidden layer, alpha=$10^{-3}$, a maximum of 300 iterations, early stopping with patience 15 and
random seed 0. O and real/permuted R are standardised using each probe's training fold.
Gaussian blocks follow the original unscaled standard-normal convention. Score probabilities
are clipped to $[10^{-6},1-10^{-6}]$. Baseline features are $(O_j,0_{L_f})$.

\subsection{R1, same-half R2 and seed replication}

R1 requires both primary p-values to be at most 0.025 within a seed, followed by at least
two positive seeds in a family--probe--view cell. The 79-spoke p-value grid has resolution
1/80; both 0.0125 and 0.025 pass. Conditional effect size is
$\Delta_{\rm cond,S}=\widehat\Delta^{\rm obs}_S-B^{-1}\sum_b\widehat\Delta^{(b)}_S$,
not an estimate certified to equal the Bayes-risk gain. A negative observed raw gain can
still exceed the conditional reference; this is not a logical contradiction.

The same-half robustness rule is
\[
D_{\rm R2,seed}=D_{\rm CPT,seed}\ \wedge\ D_{\rm noise,seed},
\qquad D_{\rm R2,cell}=\mathbf1\!\left\{\sum_{\rm seeds}D_{\rm R2,seed}\ge2\right\}.
\]
Noise positivity requires the positive-direction 95\% interval to exclude zero for both
log loss and Brier on the same $B_{\rm test}$ rows. There is no raw/global/matched conjunct.
The executed minimum-two rule retains all five available DeiT/C-100/AR-full seeds. This means two of five for that family, not a majority rule; the protocol wording
is clarified in App.~\ref{app:t2provenance}. No multiplicity-adjusted familywise
interpretation is assigned to either R1 or R2.

\subsection{Historical fold-specific primary calibration}
\label{app:histcal}

The six fixed calibration cells each contribute 100 replicates at O0/O1 and 60 at O2/O3.
Both probes are evaluated for every label record: 1,920 label records yield 3,840 two-score
replicate--probe evaluations. q fitting is independent of the 5k test half; the label
generator is the frozen output-only fold-specific recipe detailed in App.~\ref{app:exactnull}.
The calibration count and gate are empirical statements under that recipe, not a proof of
conditional independence for every data-dependent generator or of true-q CPT validity.

\begin{table}[htbp]
\centering\footnotesize
\setlength{\tabcolsep}{3pt}
\caption{Historical fold-specific controls differ in detection frequency. Counts compare identical test halves.}\label{tab:t2nullcontrols}
\begin{tabular}{llrrrr}
\toprule
View & Probe & RAW & GLOBAL & MATCHED & CPT \\
\midrule
O0 & Linear & 0/600 (0.00\%) & 2/600 (0.33\%) & 5/600 (0.83\%) & 12/600 (2.00\%) \\
O0 & MLP & 286/600 (47.67\%) & 109/600 (18.17\%) & 92/600 (15.33\%) & 13/600 (2.17\%) \\
O1 & Linear & 0/600 (0.00\%) & 3/600 (0.50\%) & 3/600 (0.50\%) & 9/600 (1.50\%) \\
O1 & MLP & 102/600 (17.00\%) & 93/600 (15.50\%) & 85/600 (14.17\%) & 14/600 (2.33\%) \\
O2 & Linear & 0/360 (0.00\%) & 3/360 (0.83\%) & 2/360 (0.56\%) & 7/360 (1.94\%) \\
O2 & MLP & 59/360 (16.39\%) & 63/360 (17.50\%) & 67/360 (18.61\%) & 6/360 (1.67\%) \\
O3 & Linear & 0/360 (0.00\%) & 4/360 (1.11\%) & 1/360 (0.28\%) & 9/360 (2.50\%) \\
O3 & MLP & 63/360 (17.50\%) & 53/360 (14.72\%) & 37/360 (10.28\%) & 4/360 (1.11\%) \\
\bottomrule
\end{tabular}
\vspace{2pt}
\begin{minipage}{\linewidth}\footnotesize
\textit{Note.} Each flag requires both proper scores. Counts are replicate--probe evaluations, not replicated family detections. The fold-specific generator qualification is in App.~\ref{app:exactnull}.
\end{minipage}
\end{table}

\begin{table}[htbp]
\centering\footnotesize
\setlength{\tabcolsep}{3pt}
\caption{Historical CPT cell rates vary within the pre-specified ceiling. All six models, four views and two probes are shown (48 cells).}\label{tab:t2nullcells}
\begin{tabular}{llrrrr}
\toprule
Mechanism/backbone/data & Probe & O0 & O1 & O2 & O3 \\
\midrule
MoE/DeiT/C10 & Linear & 1/100 (1.00\%) & 1/100 (1.00\%) & 0/60 (0.00\%) & 0/60 (0.00\%) \\
MoE/DeiT/C10 & MLP & 2/100 (2.00\%) & 1/100 (1.00\%) & 3/60 (5.00\%) & 1/60 (1.67\%) \\
ARb/Swin/C10 & Linear & 1/100 (1.00\%) & 0/100 (0.00\%) & 0/60 (0.00\%) & 0/60 (0.00\%) \\
ARb/Swin/C10 & MLP & 3/100 (3.00\%) & 2/100 (2.00\%) & 2/60 (3.33\%) & 0/60 (0.00\%) \\
ARb/DeiT/C100 & Linear & 4/100 (4.00\%) & 2/100 (2.00\%) & 1/60 (1.67\%) & 1/60 (1.67\%) \\
ARb/DeiT/C100 & MLP & 2/100 (2.00\%) & 2/100 (2.00\%) & 0/60 (0.00\%) & 2/60 (3.33\%) \\
ACT/DeiT/C100 & Linear & 4/100 (4.00\%) & 3/100 (3.00\%) & 4/60 (6.67\%) & 5/60 (8.33\%) \\
ACT/DeiT/C100 & MLP & 2/100 (2.00\%) & 3/100 (3.00\%) & 0/60 (0.00\%) & 1/60 (1.67\%) \\
ARb/ViT/C100 & Linear & 0/100 (0.00\%) & 1/100 (1.00\%) & 2/60 (3.33\%) & 2/60 (3.33\%) \\
ARb/ViT/C100 & MLP & 1/100 (1.00\%) & 2/100 (2.00\%) & 0/60 (0.00\%) & 0/60 (0.00\%) \\
ARf/DeiT/TIN & Linear & 2/100 (2.00\%) & 2/100 (2.00\%) & 0/60 (0.00\%) & 1/60 (1.67\%) \\
ARf/DeiT/TIN & MLP & 3/100 (3.00\%) & 4/100 (4.00\%) & 1/60 (1.67\%) & 0/60 (0.00\%) \\
\bottomrule
\end{tabular}
\vspace{2pt}
\begin{minipage}{\linewidth}\footnotesize
\textit{Note.} Entries are detections/evaluations (percent); training seed 0 throughout. ARb/ARf denote block/full attention residuals. The worst cell is ACT/DeiT/C100, O3/Linear: 5/60. This is historical fold-specific evidence.
\end{minipage}
\end{table}

The pooled CPT rate is 74/3,840=1.927\%. The eight pooled view/probe rates range from
4/360=1.11\% to 9/360=2.50\%. Each passes the pre-specified operational bands: 0.86--4.14\% for
600 replicates and 0.38--4.62\% for 360, with no individual cell exceeding 10\%.
The worst individual cell is ACT halting/DeiT/C100, O3/linear, 5/60=8.33\%; the best rate is
zero, attained in several cells. These non-uniform rates are not replaced by the pooled mean.

Only two of the five R1-positive families have directly corresponding calibration settings,
and those use training seed 0 only: DeiT/C100/AR-block and DeiT/TIN/AR-full. For the former,
linear O0--O3 rates are 4/100, 2/100, 1/60 and 1/60; for the latter O0/linear is 2/100.
Neither calibrates the other training seeds or the multi-seed rule. The C10 families and
C100/full have no exact mechanism/backbone/dataset counterpart. No expected number of false
families is inferred from a seed-level pooled rejection rate.

\subsection{Observed-correctness conditional evidence and diagnostics}

\begin{table}[htbp]
\centering\footnotesize
\setlength{\tabcolsep}{3pt}
\caption{R1 evidence appears in five families, while the additional same-width robustness criterion R2 remains unmet. All 12 R1-positive family--probe--view cells are shown, under the primary working model Q1; MLP cells use the original accuracy-based checkpoint selection (log-loss sensitivity: Table~\ref{tab:nllobserved}; working-model sensitivity: Table~\ref{tab:qsens}).}\label{tab:t2r1}
\begin{tabular}{lllrrrr}
\toprule
DeiT family & View & Probe & $\Delta_{\rm cond,log}$ & $\Delta_{\rm cond,Brier}$ & R1 seeds & R2 seeds \\
\midrule
C100 block & O0 & Linear & +0.003954 & +0.001460 & 3/3 & 1/3 \\
C100 block & O1 & Linear & +0.002577 & +0.000983 & 3/3 & 0/3 \\
C100 block & O2 & Linear & +0.002768 & +0.001076 & 2/3 & 0/3 \\
C100 block & O3 & Linear & +0.002926 & +0.001188 & 3/3 & 0/3 \\
C100 full & O0 & Linear & +0.001854 & +0.000636 & 2/5 & 1/5 \\
C100 full & O0 & MLP & +0.005424 & +0.001585 & 3/5 & 1/5 \\
C100 full & O2 & MLP & +0.001221 & +0.000442 & 2/5 & 0/5 \\
C10 block & O0 & Linear & +0.002462 & +0.000839 & 2/3 & 0/3 \\
C10 block & O1 & Linear & +0.001867 & +0.000758 & 2/3 & 0/3 \\
C10 block & O3 & Linear & +0.001900 & +0.000779 & 2/3 & 0/3 \\
C10 full & O0 & Linear & +0.002332 & +0.000712 & 2/3 & 0/3 \\
TIN full & O0 & Linear & +0.002423 & +0.000987 & 2/3 & 0/3 \\
\bottomrule
\end{tabular}
\vspace{2pt}
\begin{minipage}{\linewidth}\footnotesize
\textit{Note.} All families use DeiT attention-residual routing. Conditional contrasts are seed means in nats (log loss) or squared-score units (Brier), not Bayes gains. The last columns count same-seed two-score flags; R2 failure does not erase R1.
\end{minipage}
\end{table}

There are 12 R1-positive cells out of 176 and five distinct positive families out of 22.
All belong to the six DeiT attention-residual families (12/48 cells; 5/6 families); the other
16 families have 0/128 positive cells. Ten detections use linear probes and two use MLPs.
Cell means span 0.001221--0.005424 nats; the individual-seed range is wider and includes
negative effects (Table~\ref{tab:t2r1seeds}). The counts remain descriptive under dependence
between views, probes and shared evaluation images.

\label{app:cptdiagnostics}
\begin{table}[htbp]
\centering\footnotesize
\setlength{\tabcolsep}{2pt}
\caption{Working-model diagnostics vary across R1-positive settings. Ranges span available seeds for each family and output view.}\label{tab:t2q}
\begin{tabular}{llrrrrr}
\toprule
DeiT family & View & $\Delta$LL & Mahalanobis KS & PIT median KS & Acceptance & Hamming \\
\midrule
C100 block & O0 & 0.013--0.016 & 0.146--0.171 & 0.014--0.016 & 0.477--0.484 & 0.99977--0.99981 \\
C100 block & O1 & 0.079--0.093 & 0.157--0.183 & 0.014--0.015 & 0.426--0.436 & 0.99974--0.99975 \\
C100 block & O2 & 0.067--0.075 & 0.154--0.182 & 0.013--0.015 & 0.404--0.408 & 0.99969--0.99980 \\
C100 block & O3 & 0.101--0.133 & 0.177--0.188 & 0.013--0.015 & 0.321--0.326 & 0.99951--0.99971 \\
C100 full & O0 & 0.018--0.020 & 0.175--0.221 & 0.012--0.014 & 0.480--0.486 & 0.99979--0.99983 \\
C100 full & O2 & 0.048--0.077 & 0.178--0.226 & 0.013--0.015 & 0.403--0.421 & 0.99964--0.99980 \\
C10 block & O0 & 0.022--0.067 & 0.135--0.153 & 0.012--0.015 & 0.445--0.474 & 0.99976--0.99977 \\
C10 block & O1 & 0.083--0.121 & 0.141--0.160 & 0.012--0.015 & 0.397--0.424 & 0.99973--0.99977 \\
C10 block & O3 & 0.144--0.495 & 0.147--0.235 & 0.012--0.018 & 0.277--0.381 & 0.99918--0.99976 \\
C10 full & O0 & 0.031--0.038 & 0.124--0.157 & 0.015--0.016 & 0.467--0.469 & 0.99977--0.99979 \\
TIN full & O0 & 0.003--0.010 & 0.185--0.207 & 0.013--0.016 & 0.485--0.492 & 0.99978--0.99980 \\
\bottomrule
\end{tabular}
\vspace{2pt}
\begin{minipage}{\linewidth}\footnotesize
\textit{Note.} All use Q1; probe duplicates share the conditional model. No moderate/strong thresholds were pre-specified for KS, PIT, acceptance or Hamming. Diagnostics do not certify the true routing law.
\end{minipage}
\end{table}

Table~\ref{tab:t2q} gives the working-model diagnostics for every R1-positive family and view.
No R1-positive seed/view triggers the pre-specified F1 warning $\Delta\mathrm{LL}\le0$, where
$\Delta\mathrm{LL}$ is the mean held-out log-likelihood on $B_{\rm test}$ of the working model minus that of
an $O$-independent model with the same rank-normal marginal.
That does not rule out other q-model problems. At O0, the median Mahalanobis KS statistic is
0.17115 in R1-positive family/view groups versus 0.12649 elsewhere; at O2 these medians are
0.18036 and 0.14394. Their ranges overlap. In C100/full, s3/s4 supply the O0/linear and
O2/MLP rejections and have that family's two largest KS discrepancies. This is a model-fit
concern, not evidence of causality. Marginal PIT is not worse in the flagged groups and is
additionally standardised by held-out residual variability, so it is not an independent
certificate of the fitted conditional covariance.

No pre-specified moderate/strong warning thresholds exist for Mahalanobis KS, PIT, acceptance or
Hamming distance; raw values are reported without invented grades. High acceptance and
near-unit Hamming distance establish movement only.

\paragraph{Doubled-$S$ check.} The pre-specified diagnostic ($S=100$ on one cell per view; each
$p$-value should move by at most one $1/80$ step) was run on the first run in manifest order,
c10-deit-ar\_block-s0, at all four views with both probes. Its first execution called the driver
once per view, and the driver numbered the single requested view as index 0, so $O_1$--$O_3$ used
the $O_0$ seed stream and all four records stated $S=50$ although the sampler used 100. That output
is kept as a superseded diagnostic (\SHundredOldExceed{} of 16 $p$-values moved by more than one
step). After the fix, each view uses its canonical seed and records $S=100$; the recomputed $O_0$
record is identical to the first. The observed statistics reproduce the primary records to within
$2\times10^{-12}$, and every $p$-value recomputed from the 79 stored reference statistics equals the
stored one (Table~\ref{tab:doubleds}).

\begin{table}[htbp]\centering\footnotesize\setlength{\tabcolsep}{4pt}
\caption{Doubled-$S$ check on c10-deit-ar\_block-s0 with the corrected canonical view seeds: $p$-values
recomputed from the 79 stored reference statistics at $S=50$ (primary record) and $S=100$, and their
difference in grid steps of $1/80$. $^{*}$: at most $0.025$. The pre-specified criterion is a change
of at most one step for every $p$-value.}
\label{tab:doubleds}
\begin{tabular}{lllrrr}\toprule
View & Probe & Score & $p$, $S=50$ & $p$, $S=100$ & Steps \\\midrule
$O_0$ & Linear & log loss & 0.1000 & 0.1000 & 0 \\
$O_0$ & Linear & Brier & 0.1000 & 0.1125 & 1 \\
$O_0$ & MLP & log loss & 0.5750 & 0.5875 & 1 \\
$O_0$ & MLP & Brier & 0.4250 & 0.5125 & 7 \\
$O_1$ & Linear & log loss & 0.0875 & 0.2375 & 12 \\
$O_1$ & Linear & Brier & 0.0625 & 0.1250 & 5 \\
$O_1$ & MLP & log loss & 0.4375 & 0.4500 & 1 \\
$O_1$ & MLP & Brier & 0.3250 & 0.4000 & 6 \\
$O_2$ & Linear & log loss & 0.1500 & 0.1250 & 2 \\
$O_2$ & Linear & Brier & 0.1125 & 0.0750 & 3 \\
$O_2$ & MLP & log loss & 0.0125$^{*}$ & 0.0250$^{*}$ & 1 \\
$O_2$ & MLP & Brier & 0.0125$^{*}$ & 0.0250$^{*}$ & 1 \\
$O_3$ & Linear & log loss & 0.1250 & 0.1000 & 2 \\
$O_3$ & Linear & Brier & 0.1375 & 0.1000 & 3 \\
$O_3$ & MLP & log loss & 0.0375 & 0.0125$^{*}$ & 2 \\
$O_3$ & MLP & Brier & 0.0250$^{*}$ & 0.0125$^{*}$ & 1 \\
\bottomrule\end{tabular}\end{table}

\SHundredExceed{} of the 16 $p$-values move by more than one step, by up to \SHundredMaxSteps{}
steps. The \SHundredBelowFiftyWord{} $p$-values at or below $0.025$ at $S=50$ (both $O_2$/MLP
$p$-values and the $O_3$/MLP Brier $p$-value) remain so at $S=100$, and the $O_3$/MLP log-loss
$p$-value falls from 0.0375 to 0.0125, so one of the eight tests changes decision. The check does
not pass. Because the $S=100$ chains use the same seed streams, the shifts are not independent Monte
Carlo replicates; they show that $p$-values of this cell depend on the number of transition rounds
by more than the protocol allows, and they give no evidence for the adequacy of $\hat q$. The check
covers one cell per view and does not measure the effect on other tests.

\subsection{Working-model sensitivity of the R1 cells}
\label{app:qsens}

This post-hoc analysis follows \path{audit/Q_SENSITIVITY_PROTOCOL_20260925.md}, written before any
of its outcomes were computed. Its scope was fixed in advance: every training seed of the five
R1-positive families at the probe and view of each of the 12 R1 cells (42 seed tests), and, for
each such cell, the same probe and view in the Swin and ViT families with the same dataset and
routing variant (24 cells, 72 seed tests). These comparison cells are R1-negative under Q1 by the
choice of scope, so they are not a calibration set with a known null. All three working models act
on the rank-normal scores $z$ of routing: Q1 and Q2 take $z\mid O\sim\mathcal N(m(O),C)$ with a ridge
or gradient-boosted mean $m$, and Q3 takes $z\mid O\sim\mathcal N\bigl(m(O),D(O)\,C_3\,D(O)\bigr)$, where
$m$ is the selected Q1 mean, $D(O)$ is the diagonal matrix of output-dependent standard deviations
(a ridge regression of log squared residuals on $O$), and $C_3$ is a correlation matrix shared across
$O$ and fitted by Ledoit--Wolf shrinkage to the residuals scaled by $D(O)^{-1}$; $C_3$ is refitted, not
copied from Q1. Q2 and Q3 are fixed sensitivity variants: their hyperparameters are tuned by
three-fold held-out conditional log-likelihood on half A, never by labels or $p$-values, and they
are not selected in preference to Q1. Each model draws its own 79 reference assignments with $S=50$
and refits the observed statistic and all references with the original probes; the MLP keeps the
accuracy-based selection with which R1 is defined, and the selection rule is not crossed with the
working model. Q1 results are the stored primary records.

\begin{table}[htbp]\centering\footnotesize\setlength{\tabcolsep}{4pt}
\caption{Working-model sensitivity of the observed-label R1 cells. Q1 is the primary working model
(ridge mean, shared residual correlation). Q2 replaces the mean by gradient boosting; Q3 keeps the
selected Q1 mean and adds an output-dependent diagonal scale, with a shared residual correlation
refitted on the scaled residuals. Q2 and Q3 are fixed sensitivity variants, fitted on half A with
hyperparameters tuned there by three-fold held-out conditional log-likelihood; both fit less well
than Q1 in all 103 fits. Each draws its own reference assignments; MLP cells use the original
accuracy-based selection. Entries give the training seeds with both $p$-values at most $0.025$;
$^{\times}$ marks a cell that no longer meets the two-seed R1 rule. The last row counts R1 cells
(seed-level flags in parentheses) in the comparison group: the Swin and ViT families with the same
dataset and routing variant at the same probe and view, selected as R1-negative under Q1 and not a
calibration set. Post-hoc; stability does not establish the true conditional law.}
\label{tab:qsens}
\begin{tabular}{llrrrr}\toprule
Family & Probe & View & Q1 & Q2 & Q3 \\\midrule
AR-block/DeiT/C-100 & Linear & $O_0$ & 3/3 & 3/3 & 3/3 \\
AR-block/DeiT/C-100 & Linear & $O_1$ & 3/3 & 3/3 & 3/3 \\
AR-block/DeiT/C-100 & Linear & $O_2$ & 2/3 & 2/3 & 2/3 \\
AR-block/DeiT/C-100 & Linear & $O_3$ & 3/3 & 3/3 & 3/3 \\
AR-full/DeiT/C-100 & Linear & $O_0$ & 2/5 & 2/5 & 2/5 \\
AR-full/DeiT/C-100 & MLP & $O_0$ & 3/5 & 3/5 & 3/5 \\
AR-full/DeiT/C-100 & MLP & $O_2$ & 2/5 & 1/5$^{\times}$ & 1/5$^{\times}$ \\
AR-block/DeiT/C-10 & Linear & $O_0$ & 2/3 & 1/3$^{\times}$ & 1/3$^{\times}$ \\
AR-block/DeiT/C-10 & Linear & $O_1$ & 2/3 & 1/3$^{\times}$ & 1/3$^{\times}$ \\
AR-block/DeiT/C-10 & Linear & $O_3$ & 2/3 & 1/3$^{\times}$ & 1/3$^{\times}$ \\
AR-full/DeiT/C-10 & Linear & $O_0$ & 2/3 & 2/3 & 2/3 \\
AR-full/DeiT/TIN & Linear & $O_0$ & 2/3 & 2/3 & 2/3 \\
\midrule Comparison group (24 cells, 72 seed tests) & & & 0 (10) & 0 (8) & 0 (10) \\
\bottomrule\end{tabular}\end{table}

Q1 has the highest held-out log-likelihood on A in all \QUnits{} run--view fits; per example, Q2 is
lower by a median of 0.29 (minimum 0.06) and Q3 by a median of 14.1 (minimum 5.0). The median
spoke acceptance rate is \QOneAccMed{} for Q1, \QTwoAccMed{} for Q2 and \QThreeAccMed{} for Q3, with
minima \QOneAccMin{}, \QTwoAccMin{} and \QThreeAccMin{}; the minimum mean Hamming distance between
observed and reference assignments is 0.978, so every chain moves. Eight of the 12 cells keep R1
under both variants and the same four lose it under both
(Table~\ref{tab:qsens}); in each of the four, one of the two positive seeds under Q1 is no longer
positive. In the R1 cells, \QPosSeedQone{} of the \QPosSeedTests{} seed tests are flagged under Q1 and
\QPosSeedQtwo{} and \QPosSeedQthree{} under Q2 and Q3; the counts are \QNegSeedQone{}, \QNegSeedQtwo{} and \QNegSeedQthree{} of \QNegSeedTests{} in the
comparison cells, none of which reaches two seeds. Q2 and Q3 describe routing less well than Q1,
all three share the rank-normal Gaussian framework, and Q3 retains the Q1 mean. Their agreement
therefore shows that eight cells are stable to these two specified variants; it is neither an
independent test of the Q1 mean nor evidence that any of the three is the true conditional law. The 103 run--view units took 8.6 single-core hours.

\paragraph{Excluded discriminator (D3).}
The discriminator used row-wise stratified cross-validation on paired real and resampled
observations, allowing an example's output vector in a held-out row to have an
opposite-labelled counterpart in training. This invalidates the intended independent
held-out two-sample diagnostic; its AUC, accuracy and derived TV values are excluded from
the conditional-fidelity argument. It does not alter the separately computed CPT statistics,
and no post-hoc replacement discriminator is used here.

\subsection{Independent-split linear power endpoint}
\begin{table}[htbp]
\centering\footnotesize
\setlength{\tabcolsep}{3pt}
\caption{Both independent-split Linear power endpoints meet the adequacy gate. Each replicate draws 10k synthetic examples before disjoint 5k fitting and 5k testing.}\label{tab:t2power}
\begin{tabular}{lrrrrrr}
\toprule
Cell & $L_f$ & $\lambda$ & Recorded gain & Positive & Power & Gate \\
\midrule
ARb/DeiT/C100 & 22 & 0.3828815918 & 0.00993134 & 200/200 & 100\% & pass \\
ARb/Swin/C10 & 8 & 0.5706916534 & 0.00993018 & 200/200 & 100\% & pass \\
\bottomrule
\end{tabular}
\vspace{2pt}
\begin{minipage}{\linewidth}\footnotesize
\textit{Note.} Coefficients and recorded true gains were pre-specified. Counts are detections/200 replicates for each cell. These endpoints do not establish corrected MLP, full-curve or R2 power.
\end{minipage}
\end{table}

The adequacy endpoint uses 400 of the 3,000 independent-split power records of
Table~\ref{tab:power}: two fixed cells with 200 linear-probe replicates each (rep IDs 0--199) at
the 0.010-nat target. Each replay first generates all 10,000 synthetic rows using
seed $400000+\mathrm{rep}$ and the original O1/eta construction:
$T^*\sim\mathrm{Bernoulli}(\hat\eta)$ and
$R=\lambda(T^*-\hat\eta)e_1+N(0,I_{L_f})$. The stored lambda is reused. Only then is the
fixed sorted seed-204 split applied. No B-test row enters q scaling/ranks, selection or
covariance fitting; no A row enters the CPT statistic. All 400 records verify a disjoint
5k/5k partition, 79 spokes, S=50, both score statistics and the fixed deterministic seeds.

Each cell separately reaches 200/200 (100\% empirical power; Table~\ref{tab:t2power}), passing the $\ge0.80$ linear
requirement at $\le0.010$ nats. This is an empirical proportion, not certainty of population
power. No confidence interval was used as a new gate and no pooling hides a cell failure.
The independent-split endpoint does not establish the full curve, MLP power, sensitivity at every
smaller effect, or the power of a replicated CPT-plus-noise conjunction. The
observed-correctness conditional contrasts and the synthetic generator's true Bayes gains are also
different quantities; sharing nats as units does not make them interchangeable.

The historical fitted-q grid contains both linear and MLP records at n=5,000 and n=10,000,
but it fitted q on half of the tested rows and then scored all rows. Those overlapping-fit
power curves remain archived and are not primary adequacy evidence. They are neither
treated as independent-split evidence nor used to establish a corrected MLP curve.

\subsection{All same-half noise comparisons and logical completion}

The pre-specified noise-comparison set contains 28 CPT-positive seed evaluations in the 12 R1-positive cells,
all listed in Table~\ref{tab:t2noise}.
Real-routing versus padded-baseline statistics replayed the stored primary observed values
exactly for both scores in all 28 cases (maximum absolute discrepancy 0). The pre-recorded
numerical tolerance was $10^{-12}$ absolute, zero relative; it was not changed after seeing
outcomes. The baseline replay is an integrity check, not an extra detection criterion.

The Gaussian stream preserves the original draw order despite selective evaluation
(App.~\ref{app:randomstreams}). Paired example bootstraps use 2,000 resamples at seed 7.
The conditional model and stored CPT flags are reused without refitting or resampling.

Three evaluations are joint-positive: C100/block s2 at O0/linear, C100/full s3 at O0/linear,
and C100/full s4 at O0/MLP. None replicates within its cell. All other 164 cells already
have fewer than two CPT-positive seeds, so logical short-circuiting determines their R2
verdicts without computing noise. An uncomputed noise flag is stored as null, not false.
The final same-half primary result is 0/176 cells and zero distinct families. Historical
mixed-5k/10k R2 values are superseded and supply no evidence for this result.

{\footnotesize\renewcommand{\arraystretch}{0.92}\setlength{\tabcolsep}{3pt}
\begin{longtable}{lrllrrr}
\caption{Only three seed comparisons meet both noise-score conditions, in different cells. All 28 same-half comparisons in R1-positive cells are shown.}\label{tab:t2noise}\\
\toprule
DeiT family & Seed & View & Probe & Log loss $\times10^3$ & Brier $\times10^3$ & Joint \\
\midrule
\endfirsthead
\toprule
DeiT family & Seed & View & Probe & Log loss $\times10^3$ & Brier $\times10^3$ & Joint \\
\midrule
\endhead
\bottomrule
\endfoot
C10 block & 1 & O0 & Linear & \shortstack[r]{+2.466\\{}[-1.011,+6.198]} & \shortstack[r]{+0.800\\{}[-0.427,+2.054]} & no \\
C10 block & 1 & O1 & Linear & \shortstack[r]{+0.754\\{}[-2.048,+3.676]} & \shortstack[r]{+0.356\\{}[-0.703,+1.493]} & no \\
C10 block & 1 & O3 & Linear & \shortstack[r]{+1.044\\{}[-2.091,+4.077]} & \shortstack[r]{+0.428\\{}[-0.671,+1.533]} & no \\
C10 block & 2 & O0 & Linear & \shortstack[r]{-0.401\\{}[-3.385,+2.427]} & \shortstack[r]{-0.071\\{}[-1.047,+0.866]} & no \\
C10 block & 2 & O1 & Linear & \shortstack[r]{+1.256\\{}[-1.517,+3.896]} & \shortstack[r]{+0.642\\{}[-0.284,+1.587]} & no \\
C10 block & 2 & O3 & Linear & \shortstack[r]{+1.321\\{}[-1.720,+4.578]} & \shortstack[r]{+0.664\\{}[-0.330,+1.701]} & no \\
C10 full & 1 & O0 & Linear & \shortstack[r]{+2.085\\{}[-0.850,+5.033]} & \shortstack[r]{+0.715\\{}[-0.299,+1.746]} & no \\
C10 full & 2 & O0 & Linear & \shortstack[r]{+0.659\\{}[-2.560,+3.822]} & \shortstack[r]{+0.406\\{}[-0.715,+1.532]} & no \\
C100 block & 0 & O0 & Linear & \shortstack[r]{+2.725\\{}[-0.575,+6.218]} & \shortstack[r]{+0.984\\{}[-0.364,+2.370]} & no \\
C100 block & 0 & O1 & Linear & \shortstack[r]{+1.161\\{}[-1.853,+4.255]} & \shortstack[r]{+0.340\\{}[-0.916,+1.540]} & no \\
C100 block & 0 & O3 & Linear & \shortstack[r]{+1.520\\{}[-1.686,+4.847]} & \shortstack[r]{+0.754\\{}[-0.492,+2.103]} & no \\
C100 block & 1 & O0 & Linear & \shortstack[r]{+1.314\\{}[-1.749,+4.188]} & \shortstack[r]{+0.402\\{}[-0.796,+1.606]} & no \\
C100 block & 1 & O1 & Linear & \shortstack[r]{+1.055\\{}[-1.721,+3.943]} & \shortstack[r]{+0.295\\{}[-0.817,+1.408]} & no \\
C100 block & 1 & O2 & Linear & \shortstack[r]{+2.503\\{}[-0.630,+5.381]} & \shortstack[r]{+1.110\\{}[-0.091,+2.218]} & no \\
C100 block & 1 & O3 & Linear & \shortstack[r]{+2.158\\{}[-0.928,+5.145]} & \shortstack[r]{+0.901\\{}[-0.363,+2.101]} & no \\
C100 block & 2 & O0 & Linear & \shortstack[r]{+4.128\\{}[+0.346,+7.800]} & \shortstack[r]{+1.642\\{}[+0.225,+3.083]} & yes \\
C100 block & 2 & O1 & Linear & \shortstack[r]{+1.964\\{}[-1.242,+5.383]} & \shortstack[r]{+0.707\\{}[-0.527,+1.961]} & no \\
C100 block & 2 & O2 & Linear & \shortstack[r]{+3.158\\{}[-0.355,+6.533]} & \shortstack[r]{+1.102\\{}[-0.233,+2.340]} & no \\
C100 block & 2 & O3 & Linear & \shortstack[r]{+2.913\\{}[-0.489,+6.582]} & \shortstack[r]{+1.532\\{}[+0.266,+2.869]} & no \\
C100 full & 1 & O0 & MLP & \shortstack[r]{+4.770\\{}[-0.679,+10.197]} & \shortstack[r]{+1.517\\{}[-0.538,+3.493]} & no \\
C100 full & 3 & O0 & Linear & \shortstack[r]{+3.680\\{}[+0.578,+6.969]} & \shortstack[r]{+1.352\\{}[+0.080,+2.656]} & yes \\
C100 full & 3 & O0 & MLP & \shortstack[r]{+6.586\\{}[+1.063,+12.119]} & \shortstack[r]{+2.153\\{}[-0.077,+4.411]} & no \\
C100 full & 3 & O2 & MLP & \shortstack[r]{-0.601\\{}[-6.093,+4.771]} & \shortstack[r]{+0.252\\{}[-1.928,+2.365]} & no \\
C100 full & 4 & O0 & Linear & \shortstack[r]{+0.377\\{}[-2.690,+3.629]} & \shortstack[r]{+0.137\\{}[-1.129,+1.444]} & no \\
C100 full & 4 & O0 & MLP & \shortstack[r]{+9.693\\{}[+3.849,+15.139]} & \shortstack[r]{+2.847\\{}[+0.472,+5.093]} & yes \\
C100 full & 4 & O2 & MLP & \shortstack[r]{+1.423\\{}[-3.599,+6.333]} & \shortstack[r]{+0.287\\{}[-1.689,+2.274]} & no \\
TIN full & 1 & O0 & Linear & \shortstack[r]{+1.422\\{}[-1.680,+4.513]} & \shortstack[r]{+0.710\\{}[-0.477,+1.967]} & no \\
TIN full & 2 & O0 & Linear & \shortstack[r]{+2.881\\{}[-0.382,+6.155]} & \shortstack[r]{+1.309\\{}[-0.052,+2.601]} & no \\
\bottomrule
\multicolumn{7}{p{.78\linewidth}}{\textit{Note.} Score(noise) minus score(routing), mean [95\% paired-bootstrap interval], multiplied by $10^3$; 2,000 resamples, seed 7. Every listed seed is CPT-positive. Joint requires both lower endpoints above zero; all observed-statistic replay discrepancies are zero.} \\
\end{longtable}}

{\footnotesize\renewcommand{\arraystretch}{0.92}\setlength{\tabcolsep}{3pt}
\begin{longtable}{lllrrrrr}
\caption{Positive cell-level R1 findings include negative individual seed effects. All seed values for the 12 R1-positive cells are shown.}\label{tab:t2r1seeds}\\
\toprule
DeiT family & View & Probe & Seed & p (log/Brier) & $\Delta_{\rm cond,log}$ & $\Delta_{\rm cond,Brier}$ & R1 \\
\midrule
\endfirsthead
\toprule
DeiT family & View & Probe & Seed & p (log/Brier) & $\Delta_{\rm cond,log}$ & $\Delta_{\rm cond,Brier}$ & R1 \\
\midrule
\endhead
\bottomrule
\endfoot
C100 block & O0 & Linear & 0 & 0.0125/0.0125 & +0.003883 & +0.001371 & yes \\
C100 block & O0 & Linear & 1 & 0.0125/0.0125 & +0.003140 & +0.001025 & yes \\
C100 block & O0 & Linear & 2 & 0.0125/0.0125 & +0.004838 & +0.001984 & yes \\
C100 block & O1 & Linear & 0 & 0.0250/0.0250 & +0.002201 & +0.000853 & yes \\
C100 block & O1 & Linear & 1 & 0.0250/0.0125 & +0.002309 & +0.000865 & yes \\
C100 block & O1 & Linear & 2 & 0.0125/0.0125 & +0.003221 & +0.001231 & yes \\
C100 block & O2 & Linear & 0 & 0.0375/0.0375 & +0.002237 & +0.000808 & no \\
C100 block & O2 & Linear & 1 & 0.0125/0.0125 & +0.002633 & +0.001090 & yes \\
C100 block & O2 & Linear & 2 & 0.0125/0.0125 & +0.003435 & +0.001331 & yes \\
C100 block & O3 & Linear & 0 & 0.0125/0.0125 & +0.002607 & +0.001091 & yes \\
C100 block & O3 & Linear & 1 & 0.0250/0.0250 & +0.002635 & +0.001064 & yes \\
C100 block & O3 & Linear & 2 & 0.0125/0.0125 & +0.003536 & +0.001411 & yes \\
C100 full & O0 & Linear & 0 & 0.1625/0.1500 & +0.000811 & +0.000365 & no \\
C100 full & O0 & Linear & 1 & 0.0750/0.0625 & +0.001304 & +0.000538 & no \\
C100 full & O0 & Linear & 2 & 0.2750/0.5625 & +0.000605 & -0.000052 & no \\
C100 full & O0 & Linear & 3 & 0.0125/0.0125 & +0.003518 & +0.001238 & yes \\
C100 full & O0 & Linear & 4 & 0.0125/0.0125 & +0.003035 & +0.001089 & yes \\
C100 full & O0 & MLP & 0 & 0.1000/0.1500 & +0.003180 & +0.001011 & no \\
C100 full & O0 & MLP & 1 & 0.0125/0.0125 & +0.007521 & +0.002057 & yes \\
C100 full & O0 & MLP & 2 & 0.3375/0.4875 & +0.001742 & +0.000190 & no \\
C100 full & O0 & MLP & 3 & 0.0250/0.0250 & +0.006996 & +0.002371 & yes \\
C100 full & O0 & MLP & 4 & 0.0125/0.0125 & +0.007680 & +0.002294 & yes \\
C100 full & O2 & MLP & 0 & 0.8000/0.8500 & -0.002848 & -0.001041 & no \\
C100 full & O2 & MLP & 1 & 0.5125/0.4500 & +0.000410 & +0.000162 & no \\
C100 full & O2 & MLP & 2 & 0.8875/0.9000 & -0.003489 & -0.001224 & no \\
C100 full & O2 & MLP & 3 & 0.0250/0.0250 & +0.003445 & +0.001381 & yes \\
C100 full & O2 & MLP & 4 & 0.0125/0.0250 & +0.008586 & +0.002935 & yes \\
C10 block & O0 & Linear & 0 & 0.1000/0.1000 & +0.001164 & +0.000515 & no \\
C10 block & O0 & Linear & 1 & 0.0125/0.0125 & +0.004131 & +0.001347 & yes \\
C10 block & O0 & Linear & 2 & 0.0125/0.0250 & +0.002093 & +0.000656 & yes \\
C10 block & O1 & Linear & 0 & 0.0875/0.0625 & +0.000968 & +0.000476 & no \\
C10 block & O1 & Linear & 1 & 0.0125/0.0125 & +0.002759 & +0.000968 & yes \\
C10 block & O1 & Linear & 2 & 0.0250/0.0250 & +0.001873 & +0.000830 & yes \\
C10 block & O3 & Linear & 0 & 0.1250/0.1375 & +0.000913 & +0.000445 & no \\
C10 block & O3 & Linear & 1 & 0.0125/0.0125 & +0.002805 & +0.001028 & yes \\
C10 block & O3 & Linear & 2 & 0.0250/0.0250 & +0.001983 & +0.000864 & yes \\
C10 full & O0 & Linear & 0 & 0.3125/0.6625 & +0.000430 & -0.000122 & no \\
C10 full & O0 & Linear & 1 & 0.0125/0.0125 & +0.003277 & +0.001151 & yes \\
C10 full & O0 & Linear & 2 & 0.0125/0.0125 & +0.003288 & +0.001106 & yes \\
TIN full & O0 & Linear & 0 & 0.0500/0.1000 & +0.001369 & +0.000469 & no \\
TIN full & O0 & Linear & 1 & 0.0125/0.0125 & +0.002212 & +0.000918 & yes \\
TIN full & O0 & Linear & 2 & 0.0125/0.0125 & +0.003687 & +0.001574 & yes \\
\bottomrule
\multicolumn{8}{p{.96\linewidth}}{\textit{Note.} P-values are log loss/Brier. Conditional contrasts are observed gain minus mean CPT-reference gain. Negative effects are retained; these are model-relative statistics, not Bayes gains.} \\
\end{longtable}}

\begin{table}[htbp]
\centering\footnotesize
\setlength{\tabcolsep}{3pt}
\caption{Routing width varies across families but is constant across seeds and output views. The manifest covers 22 families and 68 training runs.}\label{tab:t2dimensions}
\begin{tabular}{lllrr}
\toprule
Family & Backbone & Dataset & $L_f$ & Seeds \\
\midrule
Halting / C10 & DeiT-S & CIFAR-10 & 12 & 0,1,2 \\
Halting / C100 & DeiT-S & CIFAR-100 & 12 & 0,1,2 \\
AR-block / C100 & DeiT-S & CIFAR-100 & 22 & 0,1,2 \\
AR-full / C100 & DeiT-S & CIFAR-100 & 23 & 0,1,2,3,4 \\
AR-block / C10 & DeiT-S & CIFAR-10 & 22 & 0,1,2 \\
AR-full / C10 & DeiT-S & CIFAR-10 & 23 & 0,1,2 \\
MoE / C10 & DeiT-S & CIFAR-10 & 12 & 0,1,2 \\
MoE / C100 & DeiT-S & CIFAR-100 & 12 & 0,1,2 \\
AR-block / C100 & Swin-T & CIFAR-100 & 8 & 0,1,2 \\
AR-full / C100 & Swin-T & CIFAR-100 & 16 & 0,1,2 \\
AR-block / C10 & Swin-T & CIFAR-10 & 8 & 0,1,2 \\
AR-full / C10 & Swin-T & CIFAR-10 & 16 & 0,1,2 \\
AR-block / TIN & DeiT-S & Tiny-ImageNet & 22 & 0,1,2 \\
AR-full / TIN & DeiT-S & Tiny-ImageNet & 23 & 0,1,2 \\
AR-block / TIN & Swin-T & Tiny-ImageNet & 8 & 0,1,2 \\
AR-full / TIN & Swin-T & Tiny-ImageNet & 16 & 0,1,2 \\
AR-block / TIN & ViT-B/16 & Tiny-ImageNet & 22 & 0,1,2 \\
AR-block / C100 & ViT-B/16 & CIFAR-100 & 22 & 0,1,2 \\
AR-full / C100 & ViT-B/16 & CIFAR-100 & 23 & 0,1,2 \\
AR-block / C10 & ViT-B/16 & CIFAR-10 & 22 & 0,1,2 \\
AR-full / C10 & ViT-B/16 & CIFAR-10 & 23 & 0,1,2 \\
AR-full / TIN & ViT-B/16 & Tiny-ImageNet & 23 & 0,1,2 \\
\bottomrule
\end{tabular}
\vspace{2pt}
\begin{minipage}{\linewidth}\footnotesize
\textit{Note.} Every evaluation array has 10,000 examples; $\dim(O_0)=1$, $\dim(O_1)=5$, and $\dim(O_2)=\dim(O_3)=K$ classes. Listed seeds identify training runs.
\end{minipage}
\end{table}

\FloatBarrier
\section{Local model configuration and evaluation data}
\label{app:config}
Backbone designs follow DeiT, Swin and ViT \citep{touvron2021deit,liu2021swin,dosovitskiy2021vit}.
All retained local checkpoints use the fixed final epoch (299) of 300-epoch training.
The artifact contains effective configuration snapshots for 66 of 68 canonical runs,
with a mapping to the evaluation-data registry (App.~\ref{app:provenance}); original snapshots for the two
Tiny-ImageNet ViT seed-0 runs were not found and are explicitly marked unavailable.
The per-example evaluation records and all audit calculations for those runs remain reproducible.
The snapshots specify batch sizes, effective learning rates, augmentation, optimiser,
and attention-residual block settings; these vary across historical training pipelines.
We do not describe the pooled runs as a controlled architecture-quality comparison.
All 66 available configurations enable a soft-binned calibration penalty with weight 0.1
and disable its layerwise variant. The supplied training implementation adds this term to
the supervised and architecture-specific losses, using 15 bins, soft-binning temperature
0.1 and absolute calibration gaps, following the soft-objective approach of
\citet{karandikar2021soft}. For mixup soft targets, correctness uses the target argmax.
The setting of the two unavailable historical configurations is unknown; these results
are not an audit of arbitrary cross-entropy-only training.
Attention-residual models use learned softmax weights over preceding hidden states:
full routing retains prior sublayer outputs, while block routing restricts stored states
to block boundaries. Exact block settings are recorded per available configuration.

The local MoE and soft adaptive-depth models use a DeiT-Small-shaped trunk: 12 layers,
384-dimensional embeddings, six heads, feed-forward width 1,536, and $16\times16$ patches
at resolution 224. MoE replaces every feed-forward layer with four experts and top-1
per-token routing, scales the selected expert output by its softmax gate probability,
and uses load-balancing weight 0.01. It processes every assigned token without a finite
capacity limit or token dropping. Its routing trace is the entropy of token-averaged
expert probabilities, normalised by $\log 4$, at each layer.
Soft adaptive depth executes all layers and combines their classification representations
with stick-breaking sigmoid halting weights; the final layer receives the remainder.
A ponder penalty of weight 0.01 encourages smaller expected depth. Its trace consists
of per-layer binary halting entropies, including the final computed head; it is not a
measurement of realised early-exit compute savings. Both use the same 300-epoch AdamW
recipe (weight decay 0.05, 20-epoch warmup, cosine schedule, batch 64), with exact effective
learning rates and augmentation settings in the supplied configurations.

\begin{table}[ht]\centering\footnotesize
\caption{Training quality varies across retained model families. Routing widths and top-1 accuracy ranges cover all available runs.}\label{tab:cachefacts}
\begin{tabular}{llrrr}\toprule Family & Backbone & Runs & Routing width & Accuracy range (\%)\\\midrule
Halting / C10 & DeiT-S & 3 & 12 & 91.95--92.33\\
Halting / C100 & DeiT-S & 3 & 12 & 72.13--73.37\\
AR-block / C100 & DeiT-S & 3 & 22 & 68.42--70.13\\
AR-full / C100 & DeiT-S & 5 & 23 & 70.00--72.07\\
AR-block / C10 & DeiT-S & 3 & 22 & 69.47--90.55\\
AR-full / C10 & DeiT-S & 3 & 23 & 89.72--91.80\\
MoE / C10 & DeiT-S & 3 & 12 & 91.68--92.19\\
MoE / C100 & DeiT-S & 3 & 12 & 69.37--70.33\\
AR-block / C100 & Swin-T & 3 & 8 & 72.16--72.64\\
AR-full / C100 & Swin-T & 3 & 16 & 74.99--75.80\\
AR-block / C10 & Swin-T & 3 & 8 & 90.12--90.26\\
AR-full / C10 & Swin-T & 3 & 16 & 74.44--92.39\\
AR-block / TIN & DeiT-S & 3 & 22 & 53.25--54.18\\
AR-full / TIN & DeiT-S & 3 & 23 & 53.62--54.56\\
AR-block / TIN & Swin-T & 3 & 8 & 61.22--62.14\\
AR-full / TIN & Swin-T & 3 & 16 & 63.76--64.07\\
AR-block / TIN & ViT-B/16 & 3 & 22 & 56.01--57.31\\
AR-block / C100 & ViT-B/16 & 3 & 22 & 68.81--71.73\\
AR-full / C100 & ViT-B/16 & 3 & 23 & 70.51--75.11\\
AR-block / C10 & ViT-B/16 & 3 & 22 & 73.82--89.13\\
AR-full / C10 & ViT-B/16 & 3 & 23 & 89.46--93.12\\
AR-full / TIN & ViT-B/16 & 3 & 23 & 57.77--59.39\\
\bottomrule\end{tabular}\vspace{2pt}
\begin{minipage}{\linewidth}\footnotesize
\textit{Note.} No low-accuracy run is removed. Ranges span training runs, not uncertainty intervals. Artifact-key mappings are supplied in App.~\ref{app:provenance}.
\end{minipage}
\end{table}

The attention-residual checkpoints and evaluation records are those associated with
\citet{liang2026routing}; this paper reuses them and adds audit statistics, control
comparisons, pre-specified label-regeneration stress tests, the independent-split linear power
endpoint and conditional-reference diagnostics, none of which constitutes an independently
trained replication. The MoE and soft adaptive-depth models are the local models described
above; they broaden mechanism coverage, while the statistical claims remain conditional on
the specified representations and probe recipes. Training quality varies substantially within
several families (Table~\ref{tab:cachefacts}), for example AR-block/DeiT/C-10 (69.5--90.6\%),
AR-full/Swin/C-10 (74.4--92.4\%) and AR-block/ViT/C-10 (73.8--89.1\%); no run is removed. The
lowest AR-block/DeiT/C-10 seed contributes no R1 flag; in AR-full/DeiT/C-100 the two seeds that
supply the $O_0$ and $O_2$ rejections (s3, s4) are that family's two lowest-accuracy runs
(70.0\% and 70.4\% against 70.6--72.1\%) and also the two flagged by the working-model
diagnostic (App.~\ref{app:cptdiagnostics}); we record the coincidence without reading it as a
mechanism.

\FloatBarrier
\section{Reproducibility and provenance}
\label{app:provenance}
This appendix separates computational history from the scientific evidence. Corrections
below preserve transparency about excluded records, generator limitations, fitted-power
overlap and the same-half robustness analysis; none changes the original taxonomy.

\Needspace{345pt}
\subsection{Original audit corrections}
\begin{table}[h]
\centering\footnotesize
\setlength{\tabcolsep}{4pt}
\caption{Documented corrections preserve the original decision rule and taxonomy. Entries identify excluded records, corrected inputs and analysis amendments.}
\label{tab:auditlog}
\begin{tabular}{p{0.26\linewidth}p{0.67\linewidth}}
\toprule
\textbf{Item} & \textbf{Resolution} \\
\midrule
AR-block/DeiT/C-100 classification
 & An interim reading labelled this family Type\,I on its shuffle contrast alone
   ($+0.0021$ nats, 3/3 seeds, linear). Applying the pre-specified three-way criterion literally it
   does not qualify: the raw gain is $+0.0011$ with 0/3 seeds detected, and the noise contrast
   reaches 0/3 seeds on Brier. Recorded as a partial-contrast artifact; the family is Type\,0. \\
Excluded training record
 & One nominal AR-full seed was found byte-identical in weights to the corresponding AR-block
   seed and was removed rather than repaired; an independently retrained replacement was used. \\
Repaired evaluation caches
 & Three Swin/C-100 routing caches had been overwritten; they were regenerated from the
   independent checkpoints and re-verified against their stored evaluation records. \\
Boundary cell
 & A historical $\alpha\!=\!0.05$ boundary observation in AR-full/DeiT/C-100 was adjudicated by
   two additional independent seeds ($p=0.410$, $p=0.410$); the family carries five seeds. \\
Later analysis amendments
 & Three, each recorded before the corresponding analysis was run and none changing the primary
   decision rule or taxonomy. (i) A contrast-specific power benchmark was added after the raw-gain benchmark
   (App.~\ref{app:power}). (ii) The post-hoc matched-shuffle analysis initially averaged its 20
   replicate predictions before scoring; since log loss and Brier are convex in the prediction
   this gave the control an ensemble advantage, so it was re-run scoring each replicate
   separately (App.~\ref{app:matched}). The corrected analysis is the one reported.
   (iii) A3: the three-way conjunction's operating characteristics were bounded post hoc from the
   existing benchmark; the per-replicate indicators needed for an exact joint rate were not
   retained, so Fr\'echet bounds are reported (App.~\ref{app:power}). \\
\bottomrule
\end{tabular}
\vspace{2pt}\begin{minipage}{\linewidth}\footnotesize
\textit{Note.} Historical intermediate readings are not current scientific conclusions;
the corrected evidence and remaining limitations are stated in the scientific appendices.
\end{minipage}
\end{table}

\FloatBarrier
\subsection{Conditional-analysis corrections}
\label{app:t2provenance}
The conditional protocol was fixed before its campaign. A metadata-only routing-dimension
audit corrected the common-width assumption before outcomes, preserving family-specific
$L_f$. A later audit identified the historical fitted-power overlap and mixed-sample
R2 component. The repair scope was fixed before execution as six reference gains
$\{0,.0005,.001,.002,.005,.01\}$ with 200 linear and 50 MLP replicates per cell and target,
plus 28 same-half noise comparisons, with no choice based on repair outcomes; 3,000 power
records were executed, of which 400 linear replicates at the 0.010-nat target form the adequacy
endpoint. The original 599 Phase-B artifacts and all their hashes were verified unchanged.
No primary observed-correctness CPT, primary calibration or original taxonomy result was modified.

The
protocol's repeated ``2-of-3'' wording and its five-seed manifest were inconsistent; the
repair audit fixes the inherited minimum-two aggregation without discarding seeds. This is
two of five for that family, not a majority rule.
The manifests include per-file hashes; the repair aggregate hashes sorted hash/path lines
joined by LF without a terminal LF. Repair runtime used Python 3.9.25, NumPy 1.26.4,
SciPy 1.13.1 and scikit-learn 1.6.1, with one thread per numerical library.

All 13 required caches have pre-execution hashes unchanged through repair completion.
Nine also match the older cache-provenance manifest; four later-added caches have no entry
there (C100/full s3/s4 and TIN/full s1/s2). They use the manifest paths and have exact
observed-score replays. Their older historical byte identity is not claimed where no
older hash exists. No repair execution deviation occurred; three ignored but already-hashed
log files were added in a follow-up result commit without changing any result.

The repair protocol's pre-specified interpretation matrix assigns its first outcome category
(``Outcome 1'' in the records) when the calibration gate and the linear adequacy endpoint pass and
no cell meets same-half R2, which is what occurred. The category is an operational verdict. It does
not establish validity under the true conditional law, remove the qualification of the historical
generator, repair the other diagnostics or cancel the R1 evidence in five families; it supports
only the statement that no evaluated cell meets the same-width robustness criterion.

\subsection{Common-generator computational provenance}
\label{app:t4provenance}
Record names use internal task identifiers: T4 for the common-generator benchmark, T2 and
Phase-B for the preceding conditional-analysis records, and B0 for the pre-execution contract
check, which passed 20/20 contracts. These names identify files, not scientific criteria. The
benchmark protocol's pre-specified interpretation matrix was applied as written: the conditional
test passed its calibration gate; the MLP comparisons kept inflated shuffle-control detection
rates (category ``A'' in the records); the linear comparisons had low rates for all controls
(category ``B''); and the ordering of the controls changed between settings (qualification ``E'').
The common-generator release contains 3,247 payload files and two attestations. Its
aggregate uses sorted SHA256, two spaces, repository-relative path and LF lines,
including the final LF; self-attestations are excluded from their own inventory.
All six input identities match their applicable full-path records, including original
conditional-analysis bytes where older historical manifests lack coverage.

\textit{Operational note.} No completed scientific evaluation was rerun and no protocol choice changed; full execution logs remain in the artifact.

\subsection{Deterministic random streams}
Historical artifact names and seed fields use ``rung'' for what the manuscript calls an output
view; the text uses ``view'' because the representations are not nested.
\label{app:randomstreams}
The conditional split is exactly \texttt{default\_rng(204).permutation(10000)}, with
sorted first/last 5,000 indices. The common-generator label seed is
$10^9+100{,}000\,\mathrm{cell\_idx}+\mathrm{rep}$ using the full sorted 68-tag index.
All 600 label seeds are distinct and disjoint from the control/bootstrap/CPT namespaces;
no mutable label RNG is shared with those consumers.

The primary seed algebra is
\[
5{,}000{,}000+100{,}000\,\mathrm{cell\_idx}
+10{,}000\,\mathrm{rung\_idx}+100\,\mathrm{leg}+b,
\]
where cell indices follow the sorted 68-tag manifest. Calibration and power add
$1{,}000{,}000(1+\mathrm{rep})$. Power retains $\mathrm{rung\_idx}=0$, the historical field indexing the output view, although its generator uses O1.
The conditional-model dimensionality is family-specific (Table~\ref{tab:t2dimensions}).

The original Gaussian stream at N=5,000 uses \texttt{default\_rng(1000+repeat)}. For each
fold and view, skipped training/test permutation draws are consumed before the Gaussian
block is drawn. This preserves the original stream despite selective evaluation. Paired
example bootstraps use 2,000 resamples at seed 7. q is not refit and CPT is not resampled.
The stored seed-level CPT flags are reused unchanged.

\subsection{Immutable artifact identifiers}
Tables~\ref{tab:artifactids}--\ref{tab:familykeys} record the Git commit identifiers of the result and
interpretation records, full SHA256 digests of the key protocols and result aggregates, and the
mapping from artifact family keys to manuscript labels. The following identifiers locate the complete records, including historical outcomes
that are excluded from the current evidence. Hashes identify bytes, not statistical validity.
The common-generator report is \path{audit/T4_COMMON_GENERATOR_REPORT.md}; the original
full-sample design is \path{FOLLOWUP_EXPERIMENT_PROTOCOL.md}. The complete component-power
bounds are \path{results/power/conjunction_bounds.txt}. Model and data manifests retain
machine-readable family keys; Table~\ref{tab:familykeys} maps them to the labels used here.
\begin{table}[htbp]\centering\footnotesize
\caption{Immutable records separate results from interpretation. Identifiers locate the complete artifact history.}\label{tab:artifactids}
\begin{tabularx}{\linewidth}{@{}Xlp{.18\linewidth}@{}}\toprule
Artifact / purpose & Record type & Identifier \\\midrule
Conditional results (599 artifacts) & Git commit & \texttt{f4d4e9d} \\
Conditional interpretation & Git commit & \texttt{b25e882} \\
R1 adjudication & Git commit & \texttt{c4ce715} \\
Correction specification & Git commit & \texttt{dc837c8} \\
Correction implementation & Git commit & \texttt{a1a6288} \\
Independent-split / same-half results & Git commit & \texttt{a5ff97b} \\
Original numerical result record & Git commit & \texttt{65fa31a} \\
Correction interpretation & Git commit & \texttt{3da4ab3} \\
Common-generator protocol & Git commit & \texttt{c59ac1b} \\
Protocol attestation & Git commit & \texttt{f13d981} \\
Path-validation correction & Git commit & \texttt{5c396a0} \\
Common-generator implementation & Git commit & \texttt{e92524b} \\
Common-generator results & Git commit & \texttt{d9d3688} \\
Common-generator interpretation & Git commit & \texttt{7386c4b} \\
\bottomrule\end{tabularx}
\vspace{2pt}\begin{minipage}{\linewidth}\footnotesize
\textit{Note.} Numerical records precede their interpretation reports. Hashes identify files;
they do not establish statistical validity.\end{minipage}\end{table}
\begin{table}[htbp]\centering\footnotesize
\caption{Full SHA256 identifiers support byte-level verification. Each digest is split across two lines for readability.}\label{tab:artifacthashes}
\begin{tabularx}{\linewidth}{@{}p{.38\linewidth}X@{}}\toprule
Artifact & SHA256 \\\midrule
Conditional protocol & \texttt{cca925094bf6f8094ad2bdb625ef84f2}\newline\texttt{58c67f3f7890873e42bc7166f1c2110e} \\
Conditional result aggregate & \texttt{56ffda52f85f92d937bf24d78238ff19}\newline\texttt{bf34eab5f731089d566b5e60b346cdb4} \\
Correction result aggregate & \texttt{9ae456f6e9deb9eaf0c37d01343ffe25}\newline\texttt{0bc1944ce0b72dda6fc5bbbc46de6654} \\
Common-generator protocol & \texttt{6bd38cbce3109e7ed64346adb336bf4d}\newline\texttt{37c90b395163f0690b0ba44be1b5f8cd} \\
Common-generator result aggregate & \texttt{3c23f20feb6ac518eccea0522d1c4d59}\newline\texttt{561262ce8845ca86d558c9ec93bf3442} \\
\bottomrule\end{tabularx}\end{table}

{\footnotesize
\begin{longtable}{p{.49\linewidth}p{.44\linewidth}}
\caption{Artifact family keys map to descriptive manuscript labels. Backbone and dataset retain the original manifest identity.}\label{tab:familykeys}\\
\toprule Artifact key & Manuscript identity \\\midrule\endfirsthead
\toprule Artifact key & Manuscript identity \\\midrule\endhead
\bottomrule\endfoot
\texttt{adepth\_c10} & Halting / C10 / DeiT \\
\texttt{adepth\_c100} & Halting / C100 / DeiT \\
\texttt{deit\_c100\_block} & AR-block / C100 / DeiT \\
\texttt{deit\_c100\_full} & AR-full / C100 / DeiT \\
\texttt{deit\_c10\_block} & AR-block / C10 / DeiT \\
\texttt{deit\_c10\_full} & AR-full / C10 / DeiT \\
\texttt{moe\_c10} & MoE / C10 / DeiT \\
\texttt{moe\_c100} & MoE / C100 / DeiT \\
\texttt{swin\_c100\_block} & AR-block / C100 / Swin \\
\texttt{swin\_c100\_full} & AR-full / C100 / Swin \\
\texttt{swin\_c10\_b2\_block} & AR-block / C10 / Swin \\
\texttt{swin\_c10\_full} & AR-full / C10 / Swin \\
\texttt{tin\_deit\_block} & AR-block / TIN / DeiT \\
\texttt{tin\_deit\_full} & AR-full / TIN / DeiT \\
\texttt{tin\_swin\_block} & AR-block / TIN / Swin \\
\texttt{tin\_swin\_full} & AR-full / TIN / Swin \\
\texttt{tin\_vit\_block} & AR-block / TIN / ViT \\
\texttt{vit\_c100\_block} & AR-block / C100 / ViT \\
\texttt{vit\_c100\_full} & AR-full / C100 / ViT \\
\texttt{vit\_c10\_block} & AR-block / C10 / ViT \\
\texttt{vit\_c10\_full} & AR-full / C10 / ViT \\
\texttt{vit\_tin\_full} & AR-full / TIN / ViT \\
\midrule
\path{c100-deit-ar_block-s0} & AR-block / DeiT / C100, seed 0 \\
\path{c10-swin-ar_block-s0_blocksize2} & AR-block / Swin / C10, seed 0 \\
\path{tin-deit-ar_full-s0} & AR-full / DeiT / TIN, seed 0 \\
\path{c100-vit-ar_block-s0} & AR-block / ViT / C100, seed 0 \\
\path{c10-deitmoe-s0} & MoE / DeiT / C10, seed 0 \\
\path{c100-deitad-s0} & Halting / DeiT / C100, seed 0 \\
\path{vmoe_b16_imagenet21k_randaug_strong_ft_ilsvrc2012} & V-MoE B/16 public checkpoint \\
\path{vmoe_s32_last2_ilsvrc2012_randaug_light1_ft_ilsvrc2012} & V-MoE S/32 public checkpoint \\
\end{longtable}}

\subsection{Seed audit}
\label{app:seeds}

We enumerated every random seed actually consumed by the conditional analyses, either from
the stored per-record seed lists or by reconstructing the seed algebra over the index ranges
that were executed, and searched for reuse
(\texttt{code/verify\_numbers\_and\_seeds.py}; output in
\texttt{results/seed\_audit\_20260921/}). Because $10^{6}(1+\mathrm{rep})$ equals
$10^{5}\cdot 10(1+\mathrm{rep})$, two seeds coincide exactly when they share an output view and a
$(\mathrm{leg},b)$ pair and their cell indices differ by $10(1+\Delta\mathrm{rep})$. With 68
cells this is reachable, so the aliasing is real rather than hypothetical. Three consequences
matter, and one common worry does not apply.

\paragraph{No reuse inside a single test.} Within one (cell, view) CPT, the backward leg and
the 79 spokes take 80 distinct seeds, so no permutation stream is reused inside a test. This
rules out one implementation failure mode; it is not itself evidence that the $p$-values are
valid, which rests on the conditional law and the exchangeability construction.

\paragraph{The two null benchmarks share their permutations.} Both fit the conditional model
as \texttt{fit\_tilt(O\_A, R\_A, seed\_inner=203)} on the same auxiliary rows and seed the
sampler with the same expression over the same six cells, four views and replicate range, and
the sampler depends only on $(O,R)$ and the seed. We checked this at the level of the arrays
rather than of a summary statistic. Recomputing the full $79\times5{,}000$ assignment matrix
for c100-deit-ar\_block-s0, $O_0$, replicate 0 from the stored inputs reproduces the stored
\texttt{CPT\_assignment\_sha256} exactly
(\texttt{051dc0ce963ae490\allowbreak 3cfec4f217c2f1c2\allowbreak e9302bc94f111762\allowbreak 996a055db7cfbc08}), and substituting
the other benchmark's seed expression yields a bit-identical array
(\texttt{code/verify\_shared\_permutations.py}). The second benchmark is therefore a paired
re-analysis of one permutation design under a new label law: a legitimate variance-reduction
choice, but not independent corroboration, and \S\ref{sec:results} reports only the
common-generator counts.

\paragraph{Control streams alias across views.} The control seed
$700{,}000+1{,}000\,\mathrm{cell\_idx}+7\,\mathrm{rung\_idx}+\mathrm{rep}$ repeats whenever
view index and replicate shift by $1$ and $7$. In the common-generator benchmark 558 control seeds
are shared by two to four records, so the global-shuffle assignments and Gaussian noise
blocks of those records are the same draws. Label seeds use a disjoint namespace and are
unaffected; the 600 label vectors are shared across views by design, as already reported.
The practical effect is additional dependence among pooled comparator counts beyond the
shared labels we disclose, which is a reason to read the pooled rates as descriptive
operating characteristics rather than as independent Bernoulli evidence.

\paragraph{Cross-analysis aliasing.} Observed-label CPT seeds coincide with
calibration seeds at cell indices offset by $10(1+\mathrm{rep})$, for 6{,}400 same-role
pairs. Those runs use different families, hence different $(O,R)$, so the permutations differ;
the effect is coupling of Monte Carlo error across analyses, not shared draws.

\paragraph{Status.} We did not rerun any analysis on this account. Identical integers are not
by themselves an error, and different integers would not by themselves establish
independence. What the audit changes is the reporting: the two benchmarks are paired, the
pooled comparator counts carry dependence from shared control streams as well as shared
labels, and a future execution should draw all streams from one hierarchical
\texttt{SeedSequence} spawn tree, as \texttt{cpt\_repair.py} already does for the repair
campaign.

\paragraph{Log-loss re-analysis campaign.} The corrected-policy records live in
\path{results/nll_commongen_20260921/} (\CGCov\ common-generator MLP evaluations),
\path{results/nll_observed_20260921/} (\ObsCov\ observed-label runs),
\path{results/nll_20260921/power/} (120 matched-panel records) and
\path{results/fitting_freshlabel_20260921/} (120 fixed-design records). Each record stores
the policy dictionary and the SHA-256 of the stored accuracy-policy record it is paired to, so
pairing is verified by hash rather than by filename. A read-only audit
(\path{code/audit_policy_provenance.py}, output
\path{results/policy_audit_20260921/AUDIT.json}) checks the installed policy's behaviour
against a hand-computed negative log likelihood, then per record: the policy field, input
identity by hash, assignment identity through sampler diagnostics to $10^{-12}$, the $(79,2)$
reference array and its finiteness, $p$-values on the $k/80$ grid, the presence of all three
comparators, and a producer mapping. Two limitations are recorded rather than smoothed over.
Producer identity is \emph{reconstructed} from the SLURM logs, because the records do not store
a job identifier or a code hash; and agreement of sampler diagnostics is a pairing check, not a
proof that every assignment array is bitwise identical. Array-level identity was established
separately for the two null benchmarks by recomputing the $79\times5000$ assignment array and
matching its SHA-256 to the stored value. Whether a record's continuous statistics differ from
the accuracy-policy record is a diagnostic only: two correct policies may select the same
checkpoint, so identity neither disqualifies a record nor proves the implementation is right.
The eight Tiny-ImageNet observed-label runs missing from the original allocation were computed
on 2026-09-25 with the same code and policy; each was checked against its stored reference
assignments, and the 60 earlier records are byte-identical to their previously listed hashes. The
completion run overwrote the campaign's run-metadata file. The earlier hash list, which records
the previous version of that file, is kept unchanged, a second list covers the completed campaign,
and the SLURM logs of the original allocation are kept.

\paragraph{Analyses added on 2026-09-25.} The independent PyTorch checkpoint-selection experiment
(\path{audit/TORCH_SELECTION_PROTOCOL_20260925.md}, 360 records), the working-model sensitivity
analysis (\path{audit/Q_SENSITIVITY_PROTOCOL_20260925.md}, 103 run--view units), the doubled-$S$
check (four records) and the observed-label completion above each had their protocol or scope
written before any outcome was computed. Their random streams, as coded: the PyTorch experiment
draws its validation split, weight initialisation, mini-batch order and Gaussian routing block
from its own \texttt{SeedSequence} root 20260925, and reuses the labels, implanted signals,
bootstrap indices and folds of the paired scikit-learn panel, so the two implementations are
evaluated on identical draws; the working-model sensitivity analysis seeds its reference
assignments with the primary seed algebra of this section and the canonical view index, so its
Monte Carlo draws are coupled to those of Q1; the corrected doubled-$S$ records use the primary
seeds of the same view with $S=100$; the observed-label completion uses the stored assignments.
Distinct seed integers do not by themselves make streams statistically independent. The first
doubled-$S$ execution used the $O_0$ view index for every view and recorded $S=50$; the driver was
fixed, the records are kept as superseded, and the recomputation is reported in
App.~\ref{app:cptdiagnostics}. Because the cluster queue could not start the jobs in time, the
analyses ran as low-priority processes with one thread per numerical library on a shared login
node; logs are in \path{logs_rev/}. The matched-panel columns and mean gains of
Table~\ref{tab:nllpanel} are a re-aggregation of stored per-replicate records, not new runs.

\end{document}